\documentclass[sigconf,nonacm,screen]{acmart}
\usepackage{booktabs}
\usepackage{multirow}
\usepackage{graphicx}
\usepackage[dvipsnames,table]{xcolor}
\usepackage{tabularx}
\usepackage{algorithm}
\usepackage{algpseudocode}
\usepackage{amsmath}
\usepackage{makecell}
\usepackage{balance}
\usepackage{pifont}
\usepackage[sectionbib]{bibunits}
\defaultbibliographystyle{ACM-Reference-Format}
\defaultbibliography{main}
\graphicspath{{./}{sec/}}
\newcommand{\name}{\texttt{\textbf{HeatTok}}}

\makeatletter
\let\acm@orig@fnsymbol\@fnsymbol
\renewcommand{\@fnsymbol}[1]{%
  \ifnum#1=2
    \ding{41}%
  \else
    \acm@orig@fnsymbol{#1}%
  \fi
}
\makeatother
\AtBeginDocument{%
  \renewcommand{\checkmark}{\mathord{\text{\normalfont\ding{51}}}}%
}
\AtBeginDocument{%
  }

\copyrightyear{2026}
\acmYear{2026}
\setcopyright{none}
\acmConference[MM '26]{Proceedings of the 34th ACM International Conference on Multimedia}{November 10--14, 2026}{Rio de Janeiro, Brazil}
\acmBooktitle{Proceedings of the 34th ACM International Conference on Multimedia (MM '26), November 10--14, 2026, Rio de Janeiro, Brazil}

\begin{document}

\title{HeatTok: Enhancing Remote Sensing Image Understanding via Thermodiffusion-based Tokenization}

\author{Yingying Yan}
\authornote{The first two authors contributed equally.}
\affiliation{%
  \institution{Northwestern Polytechnical University}
  \city{Xi'an}
  \country{China}
}
\email{yanyyy@mail.nwpu.edu.cn}

\author{Jiaqi Tang}
\authornotemark[1]
\affiliation{%
  \institution{Hong Kong University of Science and Technology}
  \city{Hong Kong}
  \country{China}
}
\email{jtang092@connect.ust.hk}

\author{Wei Wei}
\authornote{Corresponding author: Wei Wei.}
\affiliation{%
  \institution{Northwestern Polytechnical University}
  \city{Xi'an}
  \country{China}
}
\email{weiweinwpu@nwpu.edu.cn}

\author{Qianzhou Wang}
\affiliation{%
  \institution{Northwestern Polytechnical University}
  \city{Xi'an}
  \country{China}
}
\email{wangqianzhou@mail.nwpu.edu.cn}

\author{Jinjian Wu}
\affiliation{%
  \institution{Northwestern Polytechnical University}
  \city{Xi'an}
  \country{China}
}
\email{wujinjian@mail.nwpu.edu.cn}

\author{Botong Geng}
\affiliation{%
  \institution{Northwestern Polytechnical University}
  \city{Xi'an}
  \country{China}
}
\email{2021300094@mail.nwpu.edu.cn}

\author{Jianmin Chen}
\affiliation{%
  \institution{Northwestern Polytechnical University}
  \city{Xi'an}
  \country{China}
}
\email{chenjianmin@mail.nwpu.edu.cn}

\author{Yuyang Xia}
\affiliation{%
  \institution{Northwestern Polytechnical University}
  \city{Xi'an}
  \country{China}
}
\email{2021302800@mail.nwpu.edu.cn}

\author{Lei Zhang}
\affiliation{%
  \institution{Northwestern Polytechnical University}
  \city{Xi'an}
  \country{China}
}
\email{nwpuzhanglei@nwpu.edu.cn}

\renewcommand{\shortauthors}{Yingying Yan et al.}

\begin{abstract}
      Current visual tokenizers in Multimodal Large Language Models (MLLMs) predominantly rely on patch-based partitioning, which causes severe semantic mixture and object fragmentation in remote sensing imagery due to the irregular contours of geo-objects. Moreover, existing adaptive methods struggle to extract precise object-level tokens and lack dedicated geometric positional encodings for irregular regions. In this paper, we propose HeatTok, a semantic-aware tokenizer driven by thermodiffusion aggregation. Inspired by the physical principles of heat conduction, HeatTok adaptively merges adjacent homogeneous regions to generate semantically independent, object-aligned irregular tokens. To enable MLLMs to perceive these irregular shapes, we design the Gaussian Multimodal Rotary Positional Embedding (G-MRoPE), which models token spatial distributions via 2D Gaussians and explicitly injects center, scale, and orientation cues. Extensive evaluations on the VRSBench and EarthVQA datasets demonstrate that HeatTok effectively preserves object-level semantic integrity and achieves state-of-the-art performance under a reasonable token budget. The code is available: https://github.com/YingyingYan1/HeatTok.
\end{abstract}


\maketitle

\begin{bibunit}
\begin{figure}[t]
    \centering
    \includegraphics[width=1\linewidth]{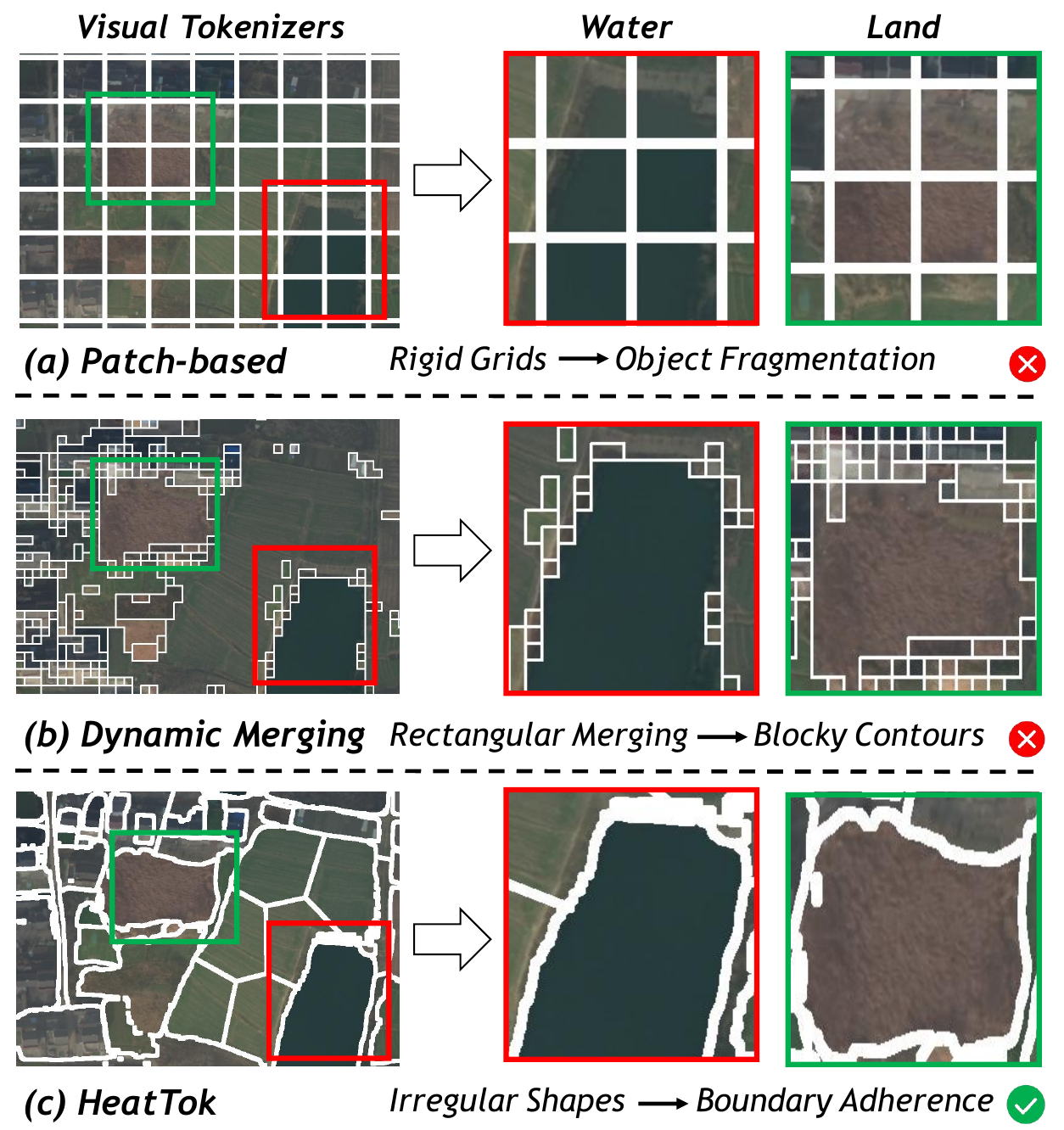}
    
    \caption{\textbf{Motivation and paradigm comparison of visual tokenizers.} Patch-based fragment irregular geo-objects, harming Semantic Integrity. Dynamic Merging groups tokens but remains grid-confined, yielding blocky outlines. HeatTok leverages thermodiffusion to generate object-level tokens that adhere to true boundaries.}
    \label{motivation}
    \end{figure}

\section{Introduction}
\label{sec:intro}

Multimodal Large Language Models (MLLMs)~\cite{yang2025qwen3,touvron2023llama,li2023blip,achiam2023gpt,li2025mini} have demonstrated significant potential in remote sensing image analysis and have been widely applied to downstream tasks such as complex scene understanding~\cite{hu2025rsgpt,kuckreja2024geochat,10655854,zhang2024earthgpt,yan2025sal,zhan2025skyeyegpt,tang2024hawk}, urban planning~\cite{li2025segearth,feng2025urbanllava,tang2026intelligent,li2024urbangpt}, and visual reasoning~\cite{wu2026iqa,tang2026robust,tang2026robustr1}. As a core component of MLLMs, the visual tokenizer is responsible for identifying basic visual elements from images and transforming them into model-processable token sequences. Its quality directly impacts the visual comprehension of the model.

Current MLLMs predominantly adopt patch-based visual tokenization~\cite{dosovitskiy2020image}, where images are uniformly partitioned into fixed-size square patches. However, this content-agnostic partitioning mechanism severely misaligns with the irregular geo-object boundaries, multi-scale targets, and complex backgrounds prevalent in remote sensing images. Regular grids fail to align with the true contours of geo-objects such as buildings and roads. This results in a mixture of multiple semantic classes within a single token, or the fragmentation of a single object across multiple tokens, as shown in Fig.~\ref{motivation}(a), thereby hindering the capability of the model to comprehend complete object structures.

To mitigate the aforementioned issues, existing studies have attempted to introduce adaptive tokenization methods, such as dynamically grouping, merging, or reallocating tokens to reduce redundancy~\cite{bolya2022token,huang2022vision,ronen2023vision}(Fig.~\ref{motivation}(b)), replacing fixed patch partitioning with adaptive sampling of positions and scales~\cite{chen2021dpt,xia2022vision,zhang2026gpstoken}, or leveraging superpixels~\cite{lew2024superpixel,aasan2024spitting,aasan2026differentiable} to generate content-aligned regions. Although these strategies enhance semantic alignment to a certain extent, they encounter two primary challenges within the context of MLLMs: First, the generated tokens predominantly remain as rectangular blocks or fragmented superpixels, failing to form irregular regions that precisely conform to complete object boundaries. Second, they lack positional encoding mechanisms specifically designed for irregular tokens in MLLMs, preventing MLLMs from effectively perceiving their geometric shapes and spatial relationships, thus limiting the practical integration of these methods.

In this paper, we propose \name{}, a semantic-aware tokenization method driven by thermodiffusion, as illustrated in Fig.~\ref{motivation}(c). \name{} first utilizes the Fast Segment Anything (FastSAM)~\cite{zhao2023fast} to generate fine-grained region proposals that tightly adhere to the boundaries of remote sensing geo-objects. To avoid the computational overhead associated with directly using these proposals, we draw inspiration from physical heat conduction principles~\cite{widder1976heat} and propose a novel ~\textbf{thermodiffusion aggregation} method. This method adaptively merges similar fine-grained regions on a region graph, elevating tokens from the ``pixel-level'' to the ``object-level.'' Consequently, it generates irregular tokens that are semantically independent, compactly shaped, and aligned with individual geo-objects. Furthermore, to resolve the geometric representation challenge of irregular tokens within MLLMs, we design a \textbf{Gaussian Multimodal Rotary Positional Embedding (G-MRoPE)}. This strategy models the spatial distribution of irregular regions via 2D Gaussian distributions~\cite{zhang2024gaussianimage}, compactly approximating the actual contours of geo-objects with elliptical shapes. By embedding the fitted Gaussian center coordinates, scale parameters, and orientation information into the Multimodal Rotary Positional Embedding (M-RoPE)~\cite{wang2024qwen2}, this approach not only provides a more robust positional representation than traditional centroids or rectangular centers but also endows the MLLM with the capability to perceive the actual coverage and geometric scale of the tokens.

To validate the effectiveness of \name{}, we integrate it into the Qwen2.5-VL~\cite{2025Qwen2} model and construct a unified evaluation framework to conduct a fair comparison with existing tokenization methods. Experimental results on VRSBench~\cite{li2024vrsbench} and EarthVQA~\cite{wang2024earthvqa} datasets, demonstrate that \name{} superiorly preserves object-level semantic integrity under a reasonable token budget. Our main contributions are summarized as follows:

\begin{itemize}

\item We propose \name{}, a semantic-aligned tokenizer that utilizes a thermodiffusion mechanism to adaptively aggregate fine-grained regions. By merging adjacent homogeneous areas, it generates boundary-adherent tokens, substantially reducing semantic mixture and fragmentation.

\item We design G-MRoPE, which injects the Gaussian center, scale, and orientation parameters of irregular tokens into M-RoPE. This endows MLLMs with robust positional representations and precise perception of scale and orientation.

\item We establish a unified evaluation benchmark on VRSBench~\cite{li2024vrsbench} and EarthVQA~\cite{wang2024earthvqa}. Extensive experiments demonstrate superior effectiveness of \name{}, achieving an accuracy of 76.37{\%} on VRSBench and 78.88{\%} on EarthVQA.

\end{itemize}

\section{ Related Work}
\label{sec:related_work}

\begin{figure*}[t]
  \centering
  \includegraphics[width=1\linewidth]{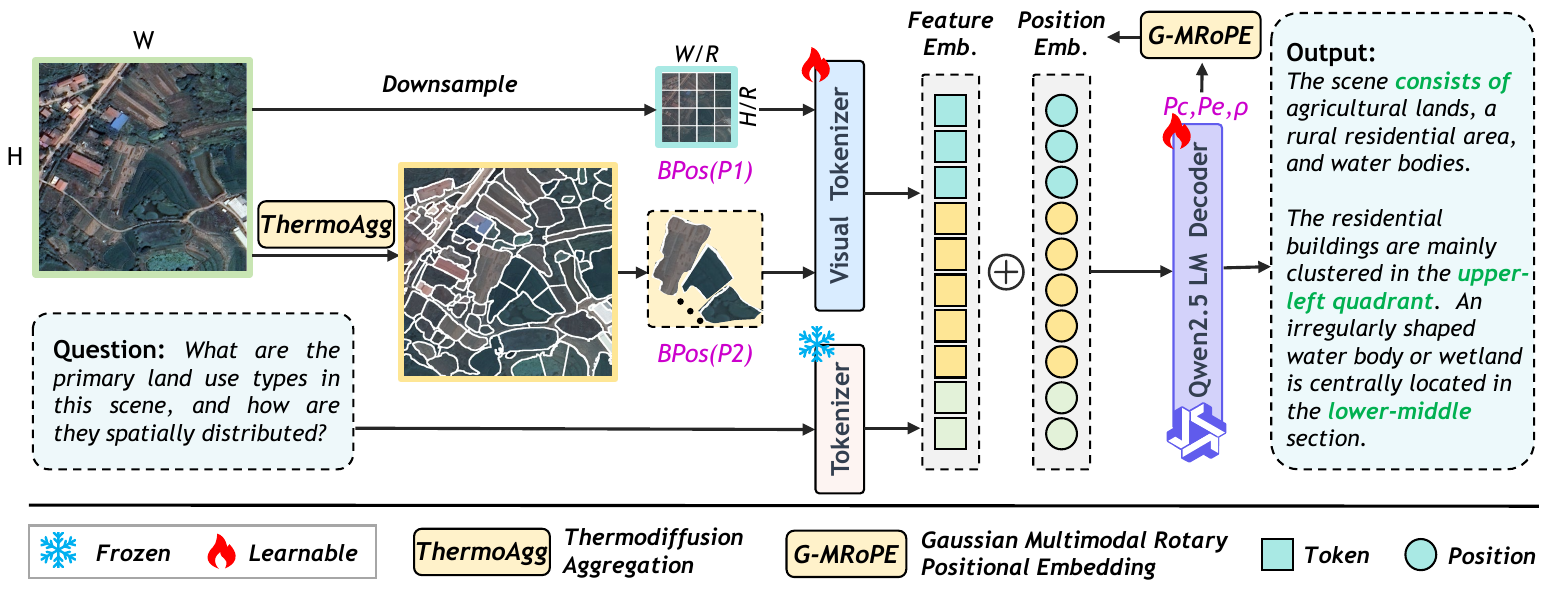}
  \caption{\textbf{The overall framework of our method.} Global branch extracts regular tokens with grid positions $BPos(P_1)$. Semantic branch uses thermodiffusion to generate object-aligned irregular tokens. Their positions $BPos(P_2)$ are parameterized into 2D Gaussians—center ($P_c$), scale ($P_e$), and orientation ($\rho$)—and explicitly injected via \textbf{G-MRoPE} before token concatenation.}
  \label{fig:framework}
\end{figure*}

Patch-based visual tokenization~\cite{dosovitskiy2020image} divides the image into square patches of fixed size. This leads to semantic fragmentation, semantic mixture within a single token, and computational redundancy. To address these issues, existing studies have begun to explore adaptive visual tokenization methods, which can be primarily categorized into three classes~\cite{aasan2026differentiable}.

\textbf{Dynamic Grouping and Merging}. These methods reduce redundancy by dynamically grouping, merging, or reallocating tokens from the original patch grid. Representative methods include ToMe~\cite{bolya2022token}, ALGM~\cite{norouzi2024algm}, Chat-UniVi~\cite{jin2024chat}, STViT~\cite{huang2022vision}, Quadtree~\cite{ronen2023vision}, HOOK~\cite{shao2024homogeneous}, and MSViT~\cite{havtorn2023msvit}. They primarily operate on the encoded grid tokens, focusing on compressing computation or adjusting the token budget based on existing patch representations, rather than redefining the visual tokenization units from the input side. Essentially, they still rely on regular patch primitives.

\textbf{Deformable Sampling}. Compared with directly merging grid tokens, this line of work replaces fixed patch partitioning with learned adaptive sampling of positions and scales. DPT~\cite{chen2021dpt} and DAT~\cite{xia2022vision} introduce deformable patch embedding and deformable attention, respectively, enabling tokens to adjust their sampling regions in a data-dependent manner. GPSToken~\cite{zhang2026gpstoken} further utilizes 2D Gaussians to parameterize each token, explicitly modeling the position and shape of the regions, thereby achieving spatially adaptive tokenization beyond uniform grids. However, the fundamental representation units of these methods remain tied to rectangular patches or the patch level. Consequently, they struggle to naturally conform to the complex boundaries of geo-objects prevalent in remote sensing imagery, such as curved rivers or tilted roads.

\textbf{Subobject and Superpixel Tokenizers}. To enhance object-level perception, researchers explore tokenization over irregular regions to align tokens with semantically consistent visual entities. Subobject methods, such as EPOC~\cite{chen2024subobject}, form adaptive regions via boundary-aware segmentation. Superpixel tokenizers, e.g., SuiT~\cite{lew2024superpixel}, $\partial $HT~\cite{aasan2026differentiable}, MSOM~\cite{dewis2026multitask}, and SPiT~\cite{aasan2024spitting}, further replace square patches with superpixel based tokens, thereby better preserving semantic integrity. However, these methods typically rely on near-pixel-level grouping strategies, thus still potentially incurring significant tokenization overhead. Furthermore, most of them employ implicit positional aggregation or simple centroid representations, lacking explicit geometric parameterization and positional encoding mechanisms tailored for irregular tokens in MLLMs.

Despite attempts to transcend rigid grids, existing adaptive tokenization strategies remain constrained by rectangular primitives or lack adequate geometric representations. Consequently, remote sensing MLLMs struggle to simultaneously preserve the semantic integrity of irregular regions~\cite{lew2024superpixel} and accurately convey their geometric morphology~\cite{zhang2026gpstoken,chen2024subobject}. To bridge this gap, we propose \name{}. By generating semantically aligned irregular tokens via thermodiffusion aggregation and integrating Gaussian parameters into M-RoPE, \name{} equips MLLMs to explicitly perceive the geometric attributes of these tokens.

\section{Methodology}

Given a remote sensing image $I \in \mathbb{R}^{H \times W \times 3}$, our objective is to generate semantically complete and geometrically rich irregular visual tokens to enhance the remote sensing image understanding of MLLMs. This problem involves two key challenges: transforming over-segmented regions into object-aligned tokens, and providing accurate geometric representations for these tokens so that MLLMs can effectively perceive their spatial extent, orientation, and relationships.

\textbf{Overall.} As illustrated in Fig.~\ref{fig:framework}, the complete pipeline of \name{} consists of two complementary branches. In the semantic branch, we first utilize thermodiffusion-guided region merging to aggregate the fine-grained proposals generated by the lightweight FastSAM~\cite{zhao2023fast} into semantically complete irregular regions, and subsequently inject their explicit geometric information into the visual tokens via G-MRoPE. Although this branch is highly effective at preserving object integrity and boundary alignment, its object-centric and spatially non-uniform sampling motivates us to retain a parallel global context branch. In this branch, the input image is downsampled by a factor of 8 to provide complementary scene-level layout and long-range spatial topology for multi-scale remote sensing imagery using only a small number of coarse-grained tokens. Finally, the tokens from both the global and semantic branches are directly concatenated to form a unified visual context for subsequent multimodal reasoning.

\subsection{Thermodiffusion-Guided Token Merging}
\label{subsec:region_merging}

\begin{figure}[t]
    \centering
    \includegraphics[width=1.0\linewidth]{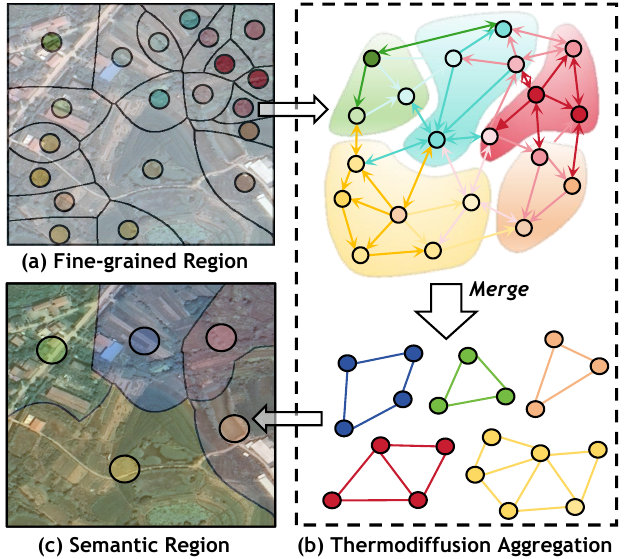}
    
    \caption{\textbf{Thermodiffusion-guided token merging.} \textbf{(a)} Initial over-segmented regions act as graph nodes. \textbf{(b)} Heat diffuses via semantic consistency to merge correlated subgraphs. \textbf{(c)} Yields semantically complete, object-aligned regions.}
    \label{fig:heat}
\end{figure}

\textit{\textbf{Preliminary.}} Thermodiffusion, also known as heat conduction, is a fundamental process in physics describing the spontaneous transfer of thermal energy from high-temperature regions to low-temperature regions. Within a two-dimensional spatial region $D \in \mathbb{R}^2$, the classic Heat Equation~\cite{widder1976heat} is formulated as a partial differential equation:
\begin{equation}
\frac{\partial u}{\partial t} = k \left( \frac{\partial^2 u}{\partial x^2} + \frac{\partial^2 u}{\partial y^2} \right) = k \nabla^2 u
\label{eq:continuous_heat}
\end{equation}

where $u(x, y, t)$ denotes temperature distribution at point $(x, y)$ at time $t$, $\nabla^2$ is the Laplacian operator, and $k$ denotes the thermal diffusivity, measuring the rate of heat transfer in a material.

Recently, this physical principle was introduced to computer vision via the vHeat architecture~\cite{wang2025building}, which models semantic propagation as adaptive thermal diffusion from patch "heat sources." Its frequency selectivity—where high-frequency components decay faster while low-frequency structures remain stable—naturally fits irregular contours of remote sensing geo-objects, addressing the semantic fragmentation caused by patch-based tokenization.

\textit{\textbf{Problem Formulation and Graph Construction.}}
Given a remote sensing image $I \in \mathbb{R}^{H \times W \times 3}$, we first utilize lightweight FastSAM~\cite{zhao2023fast} to generate initial set of fine-grained region proposals $\mathcal{S} = \{S_1, S_2, \dots, S_N\}$. As illustrated in Fig.~\ref{fig:heat}, regions generated by FastSAM typically exhibit an over-segmented state, where a single semantic object is fragmented into multiple disjoint regions.

We construct a region adjacency graph $\mathcal{G} = (\mathcal{V}, \mathcal{E})$, where vertex set $\mathcal{V} = \{v_1, v_2, \dots, v_N\}$ corresponds to FastSAM regions, and edge set $\mathcal{E}$ contains spatially adjacent region pairs $(v_i, v_j)$. Two regions $S_i$ and $S_j$ are adjacent iff they share a common boundary or their spatial distance is less than a predefined threshold $\delta$. Analogous to thermodynamics, each vertex $v_i$ is assigned a temperature $T_i$, characterizing the intensity of semantic consistency within that region; each edge $(v_i, v_j)$ is assigned a thermal diffusivity $\kappa_{ij}$, which controls the rate of heat transfer between adjacent regions.

\textit{\textbf{Calculation of Thermodynamic Parameters.}} The construction of the initial temperature field $T^{(0)} = \{T_1^{(0)}, T_2^{(0)}, \dots, T_N^{(0)}\}$ is grounded in the semantic similarity between regions. For each region $S_i$, we compute its mean feature vector $c_i \in \mathbb{R}^3$ in the LAB color space, and then define the initial temperature as
\begin{equation}
\label{eq:init_temperature}
T_i^{(0)} = \frac{1}{|N(i)|} \sum_{j \in N(i)} \exp \left( -\frac{\|c_i - c_j\|}{\sigma_T} \right)
\end{equation}
where $N(i) = \{j \mid (v_i, v_j) \in \mathcal{E}\}$ denotes the neighbors of vertex $v_i$, and $\sigma_T$ is the temperature scale. This definition ensures that vertices within homogeneous regions (i.e., those similar in color to their neighbors) acquire a higher initial temperature $T_i^{(0)} \approx 1$, acting as \textbf{Heat Sources}; vertices located at semantic boundaries (exhibiting significant color differences from their neighbors) possess temperatures approaching zero, acting as \textbf{Cold Sinks}.

\begin{figure}[t]
    \centering
    \includegraphics[width=1.0\linewidth]{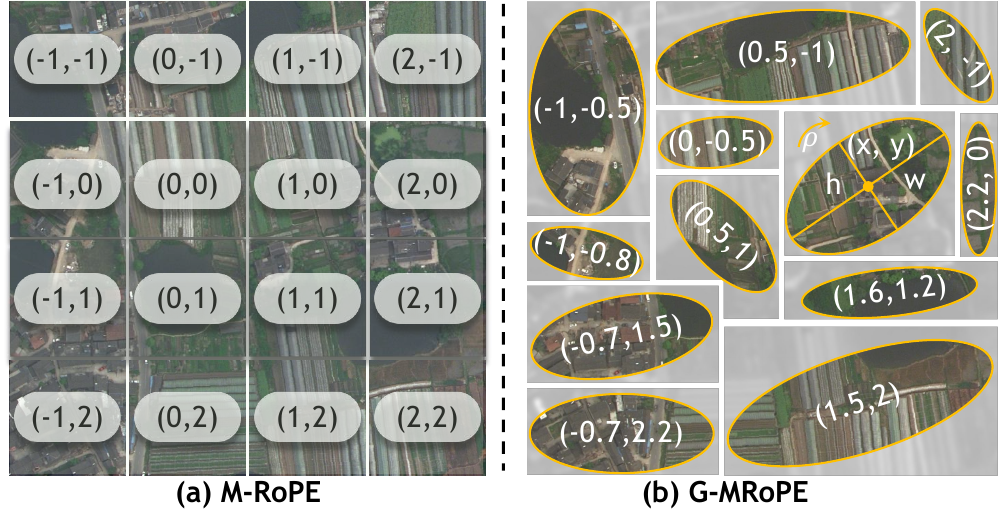}
    \caption{\textbf{(a) M-RoPE:} Assigns discrete 2D integer coordinates to uniform grids, ignoring object boundaries. \textbf{(b) G-MRoPE:} Uses 2D Gaussian fitting to explicitly encode the continuous centers $(x, y)$, scales $(w, h)$, and orientations $\rho$ of irregular regions, enabling precise perception of complex geo-objects.}
    \label{fig:MROPE}
\end{figure}

\begin{figure*}[t]
    \centering
    \includegraphics[width=1\textwidth]{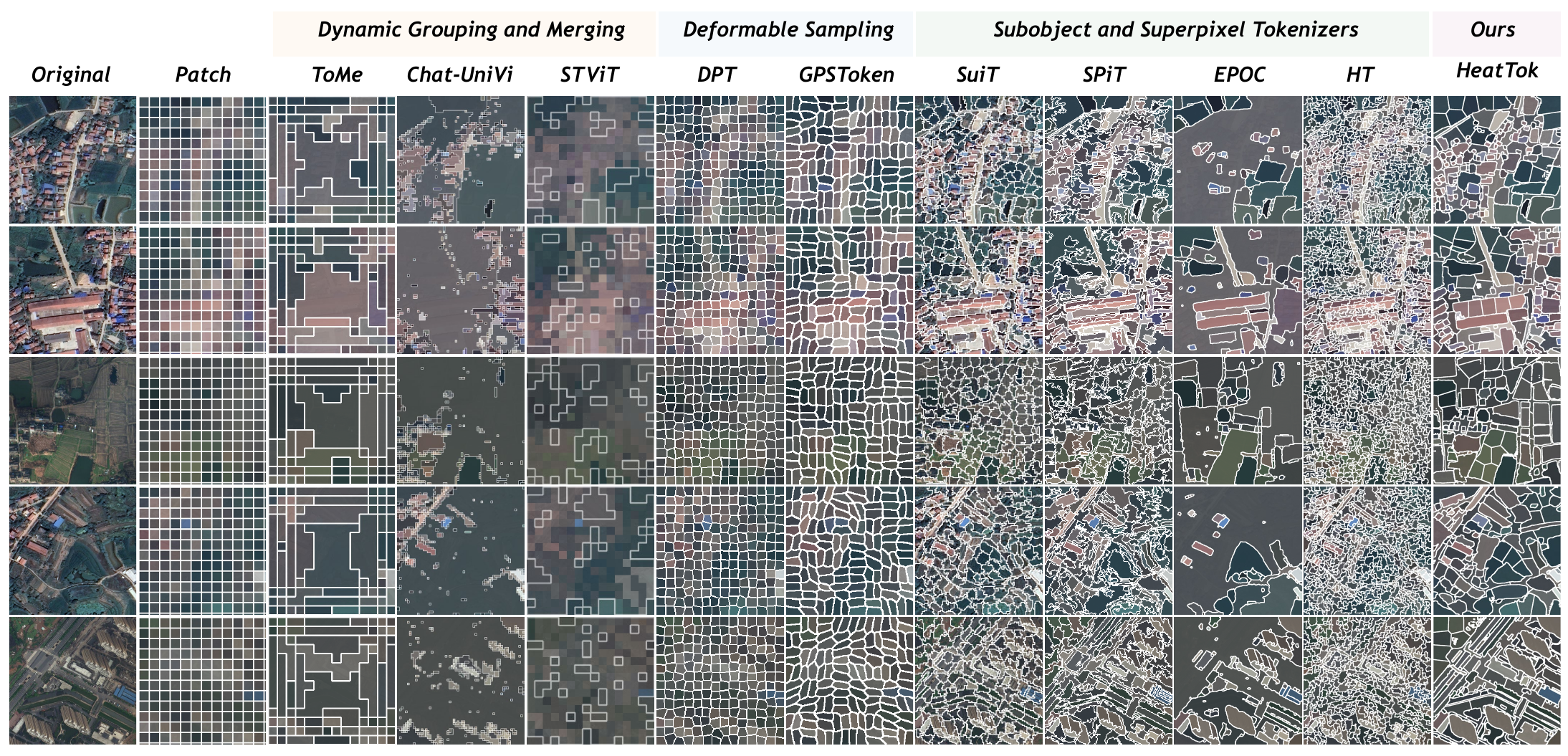}
    
    \caption{Visual comparison of semantic integrity across different tokenization methods.
    \name{} preserves crisp object boundaries and avoids over-segmentation.}
    \label{fig:vis_comparison}
\end{figure*}

The thermal diffusivity matrix $\mathbf{K} = \{\kappa_{ij}\}_{(v_i,v_j)\in\mathcal{E}}$ controls the heat transfer rate and is jointly determined by inter-region color similarity and region complexity:
\begin{equation}
\label{eq:diffusivity}
\kappa_{ij} = \kappa_0 \cdot \exp \left( -\frac{\|c_i - c_j\|_2^2}{2\sigma_c^2} \right) \cdot \left(1 + \alpha \cdot \min(C_i, C_j)\right)
\end{equation}
where $\kappa_0$ is the base diffusivity and $\sigma_c$ controls color sensitivity. The complexity $C_i = \frac{1}{|S_i|} \sum_{(x,y) \in S_i} \|\nabla I(x,y)\|$ measures average gradient over pixel set $S_i$. $\alpha \geq 0$ modulates the influence of complexity on diffusivity. This design ensures that heat diffuses faster in texture-complex regions (high $C_i$), facilitating the rapid absorption of heat source by cold sinks at boundaries, while maintaining slower diffusion in smooth regions to preserve internal stability.

\textit{\textbf{Discrete Heat Conduction Equation.}} The heat conduction process in the continuous domain is governed by the partial differential equation in Eq.~\eqref{eq:continuous_heat}. On the discrete graph structure, we employ the Graph Laplacian operator~\cite{sorkine2004laplacian} to approximate the continuous Laplacian term, yielding the iterative update formula:
\begin{equation}
\label{eq:heat_iteration}
T_i^{(t+1)} = T_i^{(t)} + \Delta t \sum_{j \in N(i)} \kappa_{ij} \left( T_j^{(t)} - T_i^{(t)} \right)
\end{equation}
where $\Delta t$ denotes the discrete time step, satisfying the stability condition $\Delta t < \frac{1}{\max_i \sum_{j \in N(i)} \kappa_{ij}}$. The iteration process proceeds until the system reaches an equilibrium state: $\max_{i \in \{1,\dots,N\}} |T_i^{(t+1)} - T_i^{(t)}| < \epsilon$, where $\epsilon$ represents a predefined convergence threshold.

\textit{\textbf{Region Merging Criteria.}}
Upon thermodiffusion convergence, we merge regions based on the final temperature field $T^{(\text{final})}$. Specifically, for each edge $(v_i, v_j) \in \mathcal{E}$, if the semantic consistency condition $|T_i^{(\text{final})} - T_j^{(\text{final})}| < \tau_m$ holds, regions $S_i$ and $S_j$ are merged into a single semantic unit. The threshold $\tau_m$ governs aggregation granularity. Following this operation, we obtain semantically complete regions $\mathcal{R} = \{R_1, R_2, \dots, R_M\}$ ($M \ll N$). Each merged region $R_k$ inherits the geometric union of its constituent regions, forming irregular shapes that tightly conform to geo-object boundaries.

\subsection{Gaussian Multimodal Rotary Positional Embedding}
\label{subsec:gaussian_pe}

\textit{\textbf{Gaussian Parameterized Modeling.}} To provide geometric representations for irregular regions $R_k \in \mathcal{R}$, traditional solutions based on bounding box centers or geometric centroids are inadequate. They exhibit spatial deviations and fail to capture scale and shape attributes, particularly for slender, curved, or non-convex geo-objects. To obtain a comprehensive geometric representation, we model each semantic region $R_k$ with a 2D Gaussian distribution $\mathcal{G}_k$. This formulation encapsulates the spatial distribution of irregular tokens into a compact five-parameter vector, characterizing their absolute centers, extents, and orientations:
\begin{equation}
\label{eq:gaussian_params}
\mathcal{G}_k = (\mu_{xk}, \mu_{yk}, \sigma_{xk}, \sigma_{yk}, \rho_k)
\end{equation}

where $\mu_{xk}, \mu_{yk} \in \mathbb{R}$ define the spatial center of the distribution, $\sigma_{xk}, \sigma_{yk} > 0$ represent the standard deviations along the horizontal and vertical axes, and the correlation coefficient $\rho_k \in [-1, 1]$ governs the orientation of the Gaussian ellipse.

Specifically, we extract the principal-axis angle from Gaussian parameters as $\varphi_k = \frac{1}{2}\operatorname{atan2}(2\rho_k\sigma_{xk}\sigma_{yk},\, \sigma_{xk}^2 - \sigma_{yk}^2)$. Here, $\varphi_k \in [-\pi/2, \pi/2]$ denotes the rotation angle of the principal axis of region $R_k$ relative to the coordinate axes. For isotropic regions ($\sigma_{xk} = \sigma_{yk}$), the orientation angle degenerates to $\varphi_k = 0$.

\begin{table*}[t]
    \centering
    \setlength{\abovecaptionskip}{2pt}
    \setlength{\belowcaptionskip}{2pt}
    \setlength{\textfloatsep}{6pt}
    \caption{Performance comparison of different tokenization methods across various dimensions on VRSBench. \textit{Tokens} denotes average number of visual tokens per image, and \textit{Throughput} denotes number of images processed per second (images/s). Red indicates the best performance, while blue denotes the second best.}
    \label{VRSBench}

    \setlength{\tabcolsep}{2.1pt}
    \renewcommand{\arraystretch}{1.1}
    \small
    
    \resizebox{0.96\textwidth}{!}{
    \begin{tabular}{c|cc|cccccccccc|c}
    \toprule
    \textbf{Visual Tokenizer} & \textbf{Tokens} & \textbf{Throughput} & \textbf{Category} & \textbf{Presence} & \textbf{Quantity} & \textbf{Color} & \textbf{Shape} & \textbf{Size} & \textbf{Position} & \textbf{Direction} & \textbf{Scene} & \textbf{Reasoning} & \textbf{All} \\
    
    \midrule
    \rowcolor[HTML]{F6F6F6} Patch Embed~\cite{dosovitskiy2020image} & 324 & 24.6 & 83.86 & 90.94 & \textcolor{red}{57.25} & 72.06 & 75.12 & 63.60 & 65.56 & 51.99 & 81.82 & 72.95 & 75.03 \\
    \midrule
    \multicolumn{14}{c}{\textbf{Dynamic Grouping and Merging}} \\
    \midrule
    ToMe~\cite{bolya2022token} & 164 & 25.43 &82.13 & \textcolor{red}{91.92} & 50.39 & 72.14 & 75.26 & 63.60 & 60.57 & 50.52 & 83.45 & 73.17 & 73.24 \\
    Chat-UniVi~\cite{jin2024chat} & 112 & 27.99 & \textcolor{red}{85.38} & \textcolor{blue}{91.50} & 55.15 & \textcolor{blue}{74.79} & 75.83 & 63.49 & \textcolor{blue}{67.57} & \textcolor{blue}{53.02} & \textcolor{blue}{83.61} & \textcolor{blue}{74.15} & \textcolor{blue}{75.87} \\
    Quadtree~\cite{ronen2023vision} & 144 & 29.97 & 56.12 & 88.01 & 50.30 & 61.52 & \textcolor{blue}{76.77} & 60.87 & 54.77 & 49.01 & 63.26 & 72.07 & 64.16 \\
    HOOK~\cite{shao2024homogeneous} & 128 & \textcolor{red}{35.33} & 70.69 & 79.33 & 60.65 & 74.85 & 73.12 & 58.73 & 59.60 & 51.17 & 79.10 & 68.41 & 69.95 \\
    STViT~\cite{huang2022vision} & 144 & 29.99 & 56.74 & 78.21 & 47.91 & 62.11 & 74.05 & 63.73 & 57.55 & 51.59 & 54.33 & 69.51 & 61.13 \\
    
    \midrule
    \multicolumn{14}{c}{\textbf{Deformable Sampling}} \\
    \midrule
    DPT~\cite{chen2021dpt} & 144 & 27.00 & 54.25 & 71.80 & 46.47 & 61.25 & 72.41 & 60.06 & 55.23 & 52.35 & 59.68 & 69.30 & 59.25\\
    GPSToken~\cite{zhang2026gpstoken} & 128 & 28.46 & 56.43 & 88.26 & 52.06 & 56.28 & 71.48 & 59.96 & 54.22 & 51.96 & 56.16 & 69.85 & 62.86 \\
    
    \midrule
    \multicolumn{14}{c}{\textbf{Subobject and Superpixel Tokenizers}} \\
    \midrule
    SuiT~\cite{lew2024superpixel} & 196 & 18.77 & 67.55 & 90.91 & 50.09 & 69.15 & 75.05 & 62.65 & 56.71 & 51.64 & 67.36 & 73.29 &  67.97\\
    SPiT~\cite{aasan2024spitting} & 196 & 18.6 & 68.44 & 91.16 & 51.59 & 65.49 & 75.19 & 63.52 & 58.31 & 52.64 & 71.33 & 72.40 & 68.82 \\
    EPOC~\cite{chen2024subobject} & $\approx 140$ & 24.64 & 64.60 & 90.67 & 49.15 & 61.29 & 74.06 & 63.82 & 51.77 & 49.38 & 70.69 & 72.29 & 66.17 \\
    $\partial $HT~\cite{aasan2026differentiable} & $\approx 150$ & 27.50 & 64.12 & 89.14 & 49.61 & 67.01 & 75.12 & \textcolor{blue}{63.91} & 56.03 & 50.64 & 68.75 & 72.18 & 66.88 \\
    \rowcolor[RGB]{224,255,255} HeatTok & $\approx 120$ & \textcolor{blue}{30.84} & \textcolor{blue}{84.45} & 91.41 & \textcolor{blue}{56.36} & \textcolor{red}{77.32} & \textcolor{red}{78.07} & \textcolor{red}{64.20} & \textcolor{red}{67.96} & \textcolor{red}{55.35} & \textcolor{red}{83.75} & \textcolor{red}{74.28} & \textcolor{red}{76.37} \\
    
    \bottomrule
    \end{tabular}
    }
    \end{table*}

    \begin{table}[t]
    \centering
    \caption{Performance comparison on the EarthVQA.}
    \label{EarthVQA}

    \scriptsize
    \setlength{\tabcolsep}{2.2pt}
    \renewcommand{\arraystretch}{1.1}
    
    \resizebox{\columnwidth}{!}{
    \begin{tabular}{c|cccccc|c}
    \toprule
    \textbf{\makecell{Visual \\ Tokenizer}} &
    \textbf{BasJu} &
    \textbf{RelJu} &
    \textbf{BasCo} &
    \textbf{RelCo} &
    \textbf{ObjAn} &
    \textbf{ComAn} &
    \textbf{All} \\
    
    \midrule
    \rowcolor[HTML]{F6F6F6} Patch Embed~\cite{dosovitskiy2020image}  & 80.44 & \textcolor{blue}{84.56} & 76.56 & 79.15 & \textcolor{blue}{66.66} & 58.15 & \textcolor{blue}{77.86} \\
    \midrule
    \multicolumn{8}{c}{\textbf{Dynamic Grouping and Merging}} \\
    \midrule
    
    ToMe~\cite{bolya2022token}  & 73.64 & 80.21 & \textcolor{red}{76.88} & 78.85 & 52.5 & 54.39 & 73.67 \\
    Chat-UniVi~\cite{jin2024chat}  & \textcolor{red}{82.74} & 80.83 & 73.68 & 77.65 & 62.09 & 55.51 & 75.72 \\
    Quadtree~\cite{ronen2023vision}  & 60.23 & 78.81 & 72.06 & 78.43 & 53.81 & 49.40 & 68.82 \\
    HOOK~\cite{shao2024homogeneous}  & 81.96 & 82.00 & 54.19 & 56.87 & 43.42 & 48.87 & 69.62 \\
    STViT~\cite{huang2022vision}  & 58.80 & 79.26 & 74.07 & 79.43 & 50.27 & 51.22 & 69.17 \\
    
    \midrule
    \multicolumn{8}{c}{\textbf{Deformable Sampling}} \\
    \midrule
    DPT~\cite{chen2021dpt}  & 62.67 & 76.35 & 70.45 & 79.43 & 50.27 & 51.76 &  68.16\\
    GPSToken~\cite{zhang2026gpstoken} & 64.81 & 75.8 & 71.91 & 78.43 & 58.11 & 50.70 & 68.94 \\
    
    \midrule
    \multicolumn{8}{c}{\textbf{Subobject and Superpixel Tokenizers}} \\
    \midrule
    SuiT~\cite{lew2024superpixel} & 80.69 & 82.12 & 73.83 & \textcolor{blue}{80.85} & 56.72 & 57.18 &  75.82\\
    SPiT~\cite{aasan2024spitting}  & 82.16 & 79.34 & 73.19 & 79.43 & 55.07 & \textcolor{blue}{58.82} &  74.98\\
    EPOC~\cite{chen2024subobject} & 64.11 & 79.65 & 73.65 & 78.43 & 53.77 & 52.96 & 70.69 \\
    $\partial$ HT~\cite{aasan2026differentiable}  & 74.88 & 80.32 & 74.08 & 78.43 & 50.27 & 53.84 & 73.18 \\
    \rowcolor[RGB]{224,255,255} \textbf{HeatTok}  & \textcolor{blue}{82.60} & \textcolor{red}{85.94} & \textcolor{blue}{75.06} & \textcolor{red}{81.67} & \textcolor{red}{68.87} & \textcolor{red}{59.49} & \textcolor{red}{78.88} \\
    \bottomrule
    \end{tabular}
    }
    \end{table}

    \begin{table*}[ht]
    \centering
    
    \caption{Case study on remote sensing visual question answering. We show the visual question answering results of a specific scenario along with the tokenization visualizations for different methods. \textcolor{red}{Red} indicates hallucinations or incorrect perceptions; \textcolor{ForestGreen}{Green} indicates accurate alignment with the ground truth.}
    \label{case}
    \small
    \begin{tabularx}{\linewidth}{X}
    \toprule
    \textbf{Remote Sensing Image \& Tokenization Visualization:} \\[1pt]
    {\centering
    \includegraphics[width=1\linewidth]{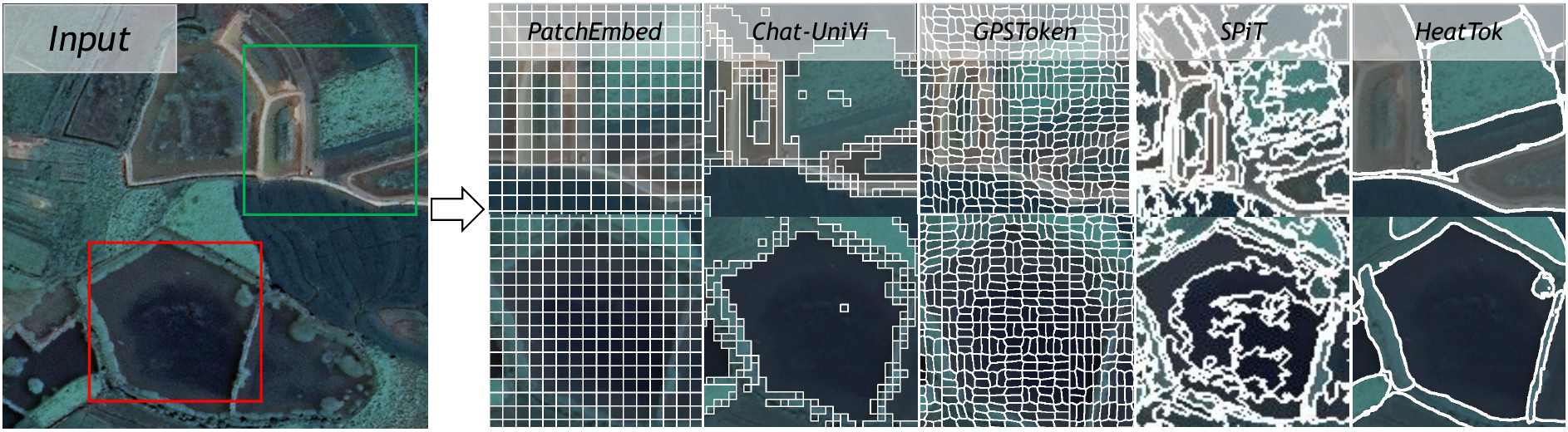}\par} \\
    
    \midrule
    \textbf{Prompt:} What are the water situations around the agricultural land? What are the needs for the renovation of villages? \\
    
    \midrule
    \textbf{Ground Truth:} There are \textcolor{ForestGreen}{polluted waters} near the agriculture land. The \textcolor{ForestGreen}{roads} need to be \textcolor{ForestGreen}{improved}, and \textcolor{ForestGreen}{waters} need to be \textcolor{ForestGreen}{cleaned up}. \\
    
    \midrule
    \textit{\textbf{Patch Embed:}} There are \textcolor{red}{no water} area. The \textcolor{ForestGreen}{roads} need to be \textcolor{ForestGreen}{improved}. \newline
    \textit{\textbf{Analysis:}} Uniform grid partitioning \textcolor{blue}{rigidly fragments} the continuous polluted water body and agricultural land. This \textcolor{blue}{disruption of semantic integrity} causes the model to fail entirely in recognizing the presence of the water body. \\
    
    \midrule
    \multicolumn{1}{c}{\textit{\textbf{Dynamic Grouping and Merging}}} \\
    \textit{\textbf{ToMe:}} There are \textcolor{red}{clean waters} near the agriculture land. \textcolor{ForestGreen}{Waters} need to be \textcolor{ForestGreen}{cleaned up}. \\[2pt]
    \textit{\textbf{Chat-UniVi:}} A \textcolor{red}{small pond} can be seen in the center of the road loop and there is a larger body of water to the upper right. The narrow dirt \textcolor{ForestGreen}{roads} between the fields need to be \textcolor{ForestGreen}{improved}, and the murky \textcolor{ForestGreen}{ponds} in the upper right need to be \textcolor{ForestGreen}{cleaned up}. \\[2pt]
    \textit{\textbf{Quadtree:}} The area seems to \textcolor{red}{lack} prominent water features near the fields. \textcolor{red}{Infrastructure} improvements, \textcolor{red}{housing} upgrades, and \textcolor{red}{agricultural} enhancements. \\[2pt]
    \textit{\textbf{HOOK:}} Farmland is completely \textcolor{red}{devoid} of nearby water resources. There are \textcolor{red}{no needs}. \\[2pt]
    \textit{\textbf{STViT:}} Two distinct \textcolor{red}{clean} water bodies are present adjacent to the fields. \textcolor{ForestGreen}{Roadways} need to be \textcolor{ForestGreen}{improved} and the nearby \textcolor{ForestGreen}{water} bodies need \textcolor{ForestGreen}{cleaning up}. \\[2pt]
    \textit{\textbf{Analysis:}} Although these methods group adjacent tokens, the Chat-UniVi visualization reveals that they remain \textcolor{blue}{constrained by rectangular primitives}. This \textcolor{blue}{blocky grouping} mixes the boundaries of the polluted water with the surrounding land, leading the model to \textcolor{blue}{hallucinate incorrect attributes}. \\
    \midrule
    \multicolumn{1}{c}{\textit{\textbf{Deformable Sampling}}} \\
    \textit{\textbf{DPT:}} There is absolutely \textcolor{red}{no water} nor any \textcolor{red}{agricultural} fields visible. There are \textcolor{red}{no needs} for the renovation of villages in this image. \newline
    \textit{\textbf{GPSToken:}} I \textcolor{red}{cannot detect} any \textcolor{red}{water} area around the agricultural land. The agricultural land needs to be \textcolor{red}{expanded}. \newline
    \textit{\textbf{Analysis:}} The GPSToken visualization shows that deformable sampling generates \textcolor{blue}{dense, spatially non-uniform cells} that capture local textures but \textcolor{blue}{disregard the macroscopic semantic structure}. Consequently, the MLLM \textcolor{blue}{loses the holistic concept} of the "polluted water," resulting in severe hallucinations and failure to recognize the renovation needs. \\
    
    \midrule
    \multicolumn{1}{c}{\textit{\textbf{Subobject and Superpixel Tokenizers}}}\\
    \textit{\textbf{SuiT:}} There are water bodies adjacent to the farm plots. There is a need to \textcolor{ForestGreen}{improve} the dirt \textcolor{ForestGreen}{roads}. \newline
    \textit{\textbf{SPiT:}} I see \textcolor{red}{clean waters} around the agricultural land. The \textcolor{ForestGreen}{roads} need \textcolor{ForestGreen}{improvement} and \textcolor{red}{housing} needs \textcolor{red}{upgrading}. \newline
    \textit{\textbf{EPOC:}} I see \textcolor{red}{no water} bodies in the area. The \textcolor{ForestGreen}{roads} need to be \textcolor{ForestGreen}{improved}. \newline
    \textit{\textbf{$\partial$HT:}} Yes, water resources are located nearby. The \textcolor{ForestGreen}{roads} need to be \textcolor{ForestGreen}{improved}. \newline
    \textit{\textbf{Analysis:}} Superpixel methods like SPiT rely on low-level pixel features, leading to \textcolor{blue}{severe over-segmentation}. The polluted water body is shattered into \textcolor{blue}{disjointed, jagged fragments}. This fragmentation makes it difficult for the model to perceive the pollution status correctly, causing it to incorrectly guess "clean waters" or miss the water cleanup requirement. \\
    
    \midrule
    \textit{\textbf{HeatTok:}} I observe that \textcolor{ForestGreen}{polluted waters} are indeed found near the agriculture land. \textcolor{ForestGreen}{Waters} need to be \textcolor{ForestGreen}{cleaned up}, and the \textcolor{ForestGreen}{roads} need to be \textcolor{ForestGreen}{improved}. \newline
    \textit{\textbf{Analysis:}} By leveraging thermodiffusion aggregation, \name{} \textcolor{blue}{more accurately outlines} the continuous, irregular boundaries of the dark water body and the agricultural plots. Preserving this \textcolor{blue}{object-level semantic integrity} helps the MLLM better ground the visual features, allowing it to identify the "polluted" status and provide more reliable renovation needs with \textcolor{blue}{fewer hallucinations}. \\
    
    \bottomrule
    \end{tabularx}
    \end{table*}

    \begin{table*}[t]
    \centering
    
    \caption{Ablation on core components of \name{}. $P^{(c)}$, $P^{(e)}$, and $\rho$ denote the Gaussian center, scale, and orientation encodings.}
    \label{tab:ablation_study}
    
    \setlength{\tabcolsep}{2.7pt}
    \renewcommand{\arraystretch}{1.05}
    \setlength{\aboverulesep}{0pt}
    \setlength{\belowrulesep}{0pt}
    
    \resizebox{\textwidth}{!}{%
    \begin{tabular}{c|ccc|cccccccccc|c}
    \toprule
    \multirow{2}{*}{\textbf{Thermodiffusion}} & \multicolumn{3}{c|}{\textbf{GMrope}} & \multirow{2}{*}{\textbf{Category}} & \multirow{2}{*}{\textbf{Presence}} & \multirow{2}{*}{\textbf{Quantity}} & \multirow{2}{*}{\textbf{Color}} & \multirow{2}{*}{\textbf{Shape}} & \multirow{2}{*}{\textbf{Size}} & \multirow{2}{*}{\textbf{Position}} & \multirow{2}{*}{\textbf{Direction}} & \multirow{2}{*}{\textbf{Scene}} & \multirow{2}{*}{\textbf{Reasoning}} & \multirow{2}{*}{\textbf{All}} \\
    \cmidrule{2-4}
     & \textbf{P(c)} & \textbf{P(e)} & \textbf{$\rho$} & & & & & & & & & & & \\
    \midrule
    
    $\times$ & $\times$ & $\times$ & $\times$ & 83.86 & 90.94 & \textbf{57.25} & 72.06 & 75.12 & 63.6 & 65.56 & 51.99 & 81.82 & 72.95 & 75.03 \\
    
    $\checkmark$ & $\times$ & $\times$ & $\times$ & 83.47 & 90.65 & 56.41 & 77.17 & 75.98 & 60.94 & 65.38 & 52.06 & 82.45 & 68.18 & 75.15 \\
    
    $\checkmark$ & $\checkmark$ & $\times$ & $\times$  & 83.82 & 91.14 & 55.96 & 76.92 & 76.15 & 61.72 & 67.34 & 52.94 & 82.81 & 73.63 & 75.72 \\
    
    $\checkmark$ & $\checkmark$ & $\checkmark$ & $\times$ & 84.26 & 91.20 & 56.27 & 77.28 & 76.54 & 63.34 & 67.86 & 53.24 & 83.62 & 74.26 & 76.14 \\
    
    $\checkmark$ & $\checkmark$ & $\checkmark$ & $\checkmark$ & \textbf{84.45} & \textbf{91.41} & 56.36 & \textbf{77.32} & \textbf{78.07} & \textbf{64.20} & \textbf{67.96} & \textbf{55.35} & \textbf{83.75} & \textbf{74.28} & \textbf{76.37} \\
    
    \bottomrule
    \end{tabular}%
    }
    \end{table*}

\textit{\textbf{Integration of G-MRoPE}.} To inject Gaussian geometric information into the MLLM, we extend the M-RoPE framework of Qwen2.5-VL~\cite{2025Qwen2} by designing a joint position-scale encoding strategy, named Gaussian multimodal rotational positional encoding (G-MRoPE), as compared in Fig.~\ref{fig:MROPE}. Let $D$ denote the visual token feature dimension. Standard RoPE~\cite{su2024roformer} defines inverse frequencies $\{\omega_m\}_{m=0}^{D/2-1}$ with $\omega_m = \theta^{-2m/D}$. For each frequency $\omega_m$ and region orientation angle $\varphi_k$, we define their respective 2D rotation matrices:
\begin{equation}
\label{eq:rotation_matrices}
\mathbf{R}(\omega_m) = \begin{pmatrix} \cos \omega_m & -\sin \omega_m \\ \sin \omega_m & \cos \omega_m \end{pmatrix}, \quad
\mathbf{R}(\varphi_k) = \begin{pmatrix} \cos \varphi_k & -\sin \varphi_k \\ \sin \varphi_k & \cos \varphi_k \end{pmatrix}
\end{equation}
To effectively incorporate the orientation prior along with the spatial context, we apply a joint rotational transformation to both the center vector $\boldsymbol{\mu}_k = (\mu_{xk}, \mu_{yk})^\top$ and the scale vector $\boldsymbol{\sigma}_k = (\sigma_{xk}, \sigma_{yk})^\top$:
\begin{equation}
\label{eq:rope_center_extent}
\mathbf{P}_{km}^{(c)} = \mathbf{R}(\omega_m)\,\mathbf{R}(\varphi_k)\,\boldsymbol{\mu}_k, \quad
\mathbf{P}_{km}^{(e)} = \mathbf{R}(\omega_m)\,\mathbf{R}(\varphi_k)\,\boldsymbol{\sigma}_k
\end{equation}

where $\mathbf{R}(\varphi_k)$ rotates the center and scale vectors into the principal axis frame of region $R_k$, and $\mathbf{R}(\omega_m)$ then applies frequency-dependent rotational encoding. This sequential mechanism decouples the intrinsic orientation of the irregular region from multi-frequency spatial encoding. Specifically, $\mathbf{R}(\varphi_k)$ aligns the coordinate vectors with the principal axis frame of the geo-object, reducing orientation-induced geometric variance, while $\mathbf{R}(\omega_m)$ projects the aligned coordinates into the frequency domain.

Through this joint transformation, the resulting vectors $\mathbf{P}_{km}^{(c)} \in \mathbb{R}^2$ and $\mathbf{P}_{km}^{(e)} \in \mathbb{R}^2$ comprehensively encode the absolute spatial coordinates and the region coverage extent, respectively. To balance the contributions of position and scale information across the limited feature dimensions, we design frequency-dependent weight coefficients defined as $\alpha_m = 1 - \frac{m}{D/2 - 1}$ and $\beta_m = 1 - \alpha_m$. Low-frequency components (small $m$) emphasize absolute position, while high-frequency components (large $m$) prioritize fine-grained scale, since lower frequencies smoothly capture stable, global geometric attributes, whereas higher frequencies are more sensitive to local spatial variations. The final MRoPE embedding is obtained through weighted combination:
\begin{equation}
\label{eq:mrope_embedding}
\begin{aligned}
E_{k,m}^{\cos} &= \alpha_m \cos(\mathbf{P}_{km}^{(c)}) + \beta_m \cos(\mathbf{P}_{km}^{(e)}) \\
E_{k,m}^{\sin} &= \alpha_m \sin(\mathbf{P}_{km}^{(c)}) + \beta_m \sin(\mathbf{P}_{km}^{(e)})
\end{aligned}
\end{equation}

This encoding is integrated with the attention mechanism of the Large Language Model backbone in Qwen2.5-VL. For the $m$-th feature slice of the query (or key) vector $\mathbf{q}_{k,m}$ associated with token $k$, position awareness is achieved via the rotary operation:

\begin{equation}
\tilde{\mathbf{q}}_{k,m} = \mathbf{q}_{k,m} \odot E_{k,m}^{\cos} + \mathbf{q}_{k,m}^\perp \odot E_{k,m}^{\sin}
\end{equation}
where $\mathbf{q}_{k,m}^\perp$ denotes adjacent-dimension swapping with sign negation of $\mathbf{q}_{k,m}$. This allows the model to preserve original geometric information when processing normalized patches, enhancing its ability to understand and localize irregular geo-objects.

\section{Experiments}
\label{sec:experiments}

\subsection{ExperimentalSettings}
\label{subsec:experimental_setup}

\textit{\textbf{Benchmarks and Metric.}} We evaluate on VRSBench~\cite{li2024vrsbench} and EarthVQA~\cite{wang2024earthvqa}, comprehensive benchmarks designed for remote sensing visual question answering. Both datasets feature diverse question types covering multiple semantic levels, ranging from basic object attributes and counting to complex scene understanding and reasoning. We use overall answer accuracy as the primary evaluation metric, and additionally report the number of visual tokens and throughput to provide a concise assessment of efficiency.

\textit{\textbf{Implementation details.}} \name{} is built upon Qwen2.5-VL-7B~\cite{2025Qwen2}. We use LoRA~\cite{hu2022lora} with a rank of 64 to fine-tune all compared methods on the VRSBench and EarthVQA datasets under the same supervised setting. Training uses a global batch size of 32, the AdamW optimizer, an initial learning rate of $1 \times 10^{-4}$ with cosine decay, and a 10{\%} linear warmup. For the Thermodiffusion module, hyperparameters are empirically set as follows: base diffusivity $\kappa_0 = 1.0$, complexity factor $\alpha = 0.5$, temperature scale $\sigma_T = 5.0$, color sensitivity scale $\sigma_C = 5.0$, and merging threshold $\tau_m = 0.03$. The diffusion iteration terminates when the maximum temperature change falls below $1 \times 10^{-5}$.

\textit{\textbf{Baselines.}} To ensure fair comparison, we establish a unified evaluation framework based on Qwen2.5-VL~\cite{2025Qwen2} by integrating compared tokenization methods into the same MLLM backbone. Specifically, we replace the Patch Embed tokenizer of Qwen2.5-VL with each baseline tokenizer while keeping the backbone and training protocol unchanged. For methods with explicit positional or geometric encodings (e.g., EPOC~\cite{chen2024subobject}, GPSToken~\cite{zhang2026gpstoken}), we incorporate their position-related information together with the M-RoPE interface of Qwen2.5-VL. For irregular-region tokenizers without explicit positional designs, we use the geometric centroid of each token region as its position to establish spatial awareness. All methods are configured to produce a comparable number of visual tokens for fair evaluation. We benchmark \name{} against representative visual tokenization methods across three paradigms: (i) Dynamic Grouping and Merging (ToMe~\cite{bolya2022token}, Chat-UniVi~\cite{jin2024chat}, Quadtree~\cite{ronen2023vision}, HOOK~\cite{shao2024homogeneous}, STViT~\cite{huang2022vision}); (ii) Deformable Sampling (DPT~\cite{chen2021dpt}, GPSToken~\cite{zhang2026gpstoken}); and (iii) Subobject and Superpixel Tokenizers (SuiT~\cite{lew2024superpixel}, SPiT~\cite{aasan2024spitting}, EPOC~\cite{chen2024subobject}, $\partial $HT~\cite{aasan2026differentiable}).

\subsection{Main results}
\label{subsec:main_results}

\textit{\textbf{Semantic Integrity of HeatTok}.}
Fig.~\ref{fig:vis_comparison} visualizes tokenization results to demonstrate semantic integrity of \name{} against three paradigms. Dynamic grouping methods (e.g., ToMe, Chat-UniVi) remain constrained by rigid rectangular grids, producing jagged contours that miss object boundaries. Deformable sampling (e.g., GPSToken) generates localized cells that often disregard semantic structures. Furthermore, subobject and superpixel tokenizers (e.g., SuiT, SPiT, $\partial $HT) rely on low-level pixel grouping, leading to over-segmentation and shattered entities. In contrast, \name{} leverages thermodiffusion to \textbf{aggregate} homogeneous regions. It merges over-segmented parts into compact, semantically coherent tokens that adhere to crisp boundaries. This enables \name{} to achieve a superior \textbf{object-level} representation, accurately capturing the semantic structures of complex geo-objects.

\textit{\textbf{Comparison with Other Tokenization Methods}.}
Tables~\ref{VRSBench} and~\ref{EarthVQA} report performance on VRSBench and EarthVQA. \name{} achieves the best overall performance on both datasets, reaching 76.37\% on VRSBench and 78.88\% on EarthVQA, while maintaining a competitive token budget and favorable throughput. In contrast, other semantic tokenization methods (e.g., DPT, GPSToken, and EPOC) perform worse on VRSBench. This suggests that these tokenizers are less suitable for remote sensing understanding, as they struggle to preserve complete object-level semantics of irregular geo-objects. Moreover, without positional encodings designed for irregular regions and compatible with M-RoPE, models may lose critical geometric details, hindering accurate visual-linguistic alignment in MLLMs. Analysis shows that \name{} is particularly effective on \textbf{Shape}, \textbf{Size}, \textbf{Position}, and \textbf{Direction}, suggesting that G-MRoPE helps the model better capture the spatial extent, morphology, and orientation of irregular regions.

\textit{\textbf{Case Study}.}
To demonstrate \name{}'s superiority in remote sensing visual question answering, Table~\ref{case} compares tokenization methods on a scene involving agricultural land, polluted water, and village renovation needs. Existing methods exhibit recognition errors and hallucinations. Patch Embed fails to recognize the water body, predicting \textit{"no water"}. Dynamic grouping and merging methods, such as ToMe and STViT, misclassify polluted water as \textit{"clean waters"}, while HOOK outputs \textit{"no needs"}. Deformable sampling methods, including DPT and GPSToken, hallucinate \textit{"no water"}, \textit{"no needs"}, and \textit{"expanded"} agricultural land. Superpixel-based methods fragment the polluted water body, weakening pollution and cleanup recognition. In contrast, \name{} captures spatial relations among the irregular water body, farmland, and roads, correctly identifying \textit{polluted waters} and renovation needs.

\subsection{Ablation Study}
\label{subsec:ablation_study}

To verify the effectiveness of core components in \name{}, we conducted ablation studies on VRSBench. As shown in Table~\ref{tab:ablation_study}, using Qwen2.5-VL Patch Embed as baseline (Row 1), we incorporate thermodiffusion aggregation, Gaussian center encoding ($P^{(c)}$), Gaussian scale encoding ($P^{(e)}$), and Gaussian orientation encoding ($\rho$).

\textit{\textbf{Effectiveness of Thermodiffusion Aggregation.}}
Introducing thermodiffusion alone (Row 2) improves semantic metrics like Color (72.06{\%} $\rightarrow$ 77.17{\%}) and Shape (75.12{\%} $\rightarrow$ 75.98{\%}), validating that forming semantically coherent regions preserves object contours. However, because irregular tokens disrupt grid-based positional structures, spatial-reliant metrics (e.g., Size, Position) slightly decline, underscoring the necessity of geometric parameterization.

\textit{\textbf{Contribution of Geometric Information in G-MRoPE.}} Building upon thermodiffusion, the incremental integration of G-MRoPE restores geometric contexts. First, embedding Gaussian center coordinates ($P^{(c)}$) resolves the spatial disorder introduced by irregular tokens. By providing a spatial anchor, this component endows the model with localization capabilities, significantly boosting \textbf{Position} accuracy from 65.38{\%} to 67.34{\%}. Subsequently, we introduce Gaussian scale encoding ($P^{(e)}$). Since irregular tokens exhibit varying spatial extents, standard point-based coordinates fail to convey object dimensions. Explicit scale parameterization equips the model with precise size perception, yielding a gain in the \textbf{Size} metric from 61.72{\%} to 63.34{\%}. Finally, incorporating the Gaussian orientation parameter ($\rho$) completes the representation. This prior captures the rotational properties of geo-objects—a crucial attribute for interpreting tilted buildings or winding roads—advancing \textbf{Direction} accuracy from 53.24{\%} to 55.35{\%}. Collectively, G-MRoPE ensures MLLMs fully comprehend the complex spatial layouts and morphologies of irregular tokens.

\section{Conclusion}

In this study, we proposed \name{} to address the critical issue of semantic fragmentation caused by Patch-based tokenization in remote sensing MLLMs. By leveraging a novel thermodiffusion aggregation mechanism, our method successfully generates irregular, object-aligned tokens that preserve the semantic integrity of complex geo-objects. Furthermore, we introduced G-MRoPE to explicitly inject Gaussian geometric priors, enabling the model to accurately perceive the precise spatial extent and location of these irregular tokens. Extensive experiments on the VRSBench and EarthVQA datasets demonstrate that \name{} better preserves object-level semantic integrity than existing methods under a reasonable token budget, achieving state-of-the-art performance.

\balance
\putbib
\end{bibunit}

\clearpage
\appendix
\begin{bibunit}

\section{Methodological Details}
\label{sec:method_details}

This section details the mathematical derivation of our thermodiffusion-guided token merging mechanism. We generalize the classical physical heat diffusion process from a continuous domain to irregular graph structures constructed by FastSAM~\cite{zhao2023fast}, thereby facilitating semantic-aware feature smoothing and aggregation.

\textit{\textbf{Theoretical Foundation: Classical Heat Equation.}}
Our merging methodology is physically inspired by the classical Heat Equation~\cite{widder1976heat}. In a two-dimensional continuous space $\Omega \subset \mathbb{R}^2$, assuming an isotropic homogeneous medium, the evolution of the temperature field $u(x, y, t)$ over time $t$ is described by the following Partial Differential Equation (PDE):
\begin{equation}
    \frac{\partial u}{\partial t} = k \left(\frac{\partial^2 u}{\partial x^2} + \frac{\partial^2 u}{\partial y^2}\right) = k \nabla^2 u
    \label{eq:supp_continuous_heat}
\end{equation}
where:
\begin{itemize}
    \item $\frac{\partial u}{\partial t}$ denotes the rate of change of temperature $u$ with respect to time.
    \item $k$ represents the thermal diffusivity, which governs the ease of feature propagation. In classical isotropic media, $k$ is a constant.
    \item $\nabla^2 u = \nabla \cdot \nabla u$ is the Laplacian operator, describing the net heat flux (diffusion) from higher temperature regions to lower temperature regions.
\end{itemize}

\textit{\textbf{Graph Construction and Discretization.}}
In our context, the visual input is not processed as a regular grid but as an \textbf{irregular graph} $\mathcal{G}=(\mathcal{V}, \mathcal{E})$ initialized by FastSAM. To achieve semantic-aware merging, we generalize the isotropic diffusion into an anisotropic process on the graph.
\begin{itemize}
    \item \textbf{Vertices $\mathcal{V}$:} Each node $v_i \in \mathcal{V}$ corresponds to a fine-grained semantic region $S_i$ generated by FastSAM. We define the semantic state of node $v_i$ as its \textbf{temperature} $T_i$.
    \item \textbf{Edges $\mathcal{E}$:} An edge $(v_i, v_j) \in \mathcal{E}$ exists if regions $S_i$ and $S_j$ are spatially adjacent. The constant diffusivity $k$ is generalized into an edge-specific weight $\kappa_{ij}$, representing the semantic affinity (thermal conductivity) between different regions.
\end{itemize}

To apply Eq.~(\ref{eq:supp_continuous_heat}) to this graph structure, we perform discretization in both temporal and spatial dimensions.

\textbf{Temporal Discretization:}
We employ Forward Euler method~\cite{euler1845institutionum} to approximate the time derivative with a discrete time step $\Delta t$:
\begin{equation}
    \frac{\partial T_i}{\partial t} \approx \frac{T_i^{(t+1)} - T_i^{(t)}}{\Delta t}
    \label{eq:supp_temporal_discrete}
\end{equation}
where $T_i^{(t)}$ denotes the temperature (feature representation) of node $i$ at iteration $t$.

\textbf{Spatial Discretization (Graph Laplacian):}
On the discrete graph, the continuous Laplacian operator $k \nabla^2 u$ is replaced by the Graph Laplacian~\cite{sorkine2004laplacian}. The net heat flux received by node $i$ is the aggregate of fluxes from all its spatial neighbors $j \in N(i)$. According to the diffusion principle, the flux from neighbor $j$ to $i$ is proportional to their temperature difference $(T_j^{(t)} - T_i^{(t)})$, dynamically weighted by the inter-region diffusivity $\kappa_{ij}$. Thus, the spatial term is discretized as:
\begin{equation}
    k \nabla^2 u \Big|_i \approx \sum_{j \in N(i)} \kappa_{ij} \cdot \left(T_j^{(t)} - T_i^{(t)}\right)
    \label{eq:supp_spatial_discrete}
\end{equation}

\begin{algorithm}[H]
    \caption{Thermodiffusion-Guided Token Merging}
    \label{alg:heat_token_merging}
    \begin{algorithmic}[1]
    \Require Input Image $I \in \mathbb{R}^{H \times W \times 3}$
    \Require FastSAM model for generating initial region proposals
    \Require Thermodynamic params: $\kappa_0, \sigma_C, \sigma_T, \alpha$
    \Require Integration params: Time step $\Delta t$, tolerance $\epsilon$
    \Require Merging threshold $\tau_m$
    
    \Statex \textit{\textcolor[RGB]{50,134,187}{// Step 1. Graph Construction and Initialization} }
    \State $\mathcal{S} \leftarrow$ Generate fine-grained region proposals $\{S_1, \dots, S_N\}$ via FastSAM
    \State $\mathcal{G}(\mathcal{V}, \mathcal{E}) \leftarrow$ Construct region adjacency graph where edges $\mathcal{E}$ represent spatial adjacency
    \For{each region $S_i \in \mathcal{S}$}
        \State $c_i, C_i \leftarrow$ Calculate LAB color mean and texture complexity
        \State $T_i^{(0)} \leftarrow$ Initialize temperature using $\sigma_T$ via Eq.~\eqref{eq:init_temperature}
    \EndFor
    \For{each edge $(v_i, v_j) \in \mathcal{E}$}
        \State $\kappa_{ij} \leftarrow$ Calculate thermal diffusivity using $(\kappa_0, \sigma_C, \alpha)$ via Eq.~\eqref{eq:diffusivity}
    \EndFor
    
    \Statex \textit{\textcolor[RGB]{50,134,187}{// Step 2. Thermodiffusion Process} }
    \State $t \leftarrow 0$
    \State $\Delta T_{max} \leftarrow \infty$
    \While{$\Delta T_{max} \geq \epsilon$}
        \For{each region $v_i \in \mathcal{V}$}
            \State $T_i^{(t+1)} \leftarrow T_i^{(t)} + \Delta t \sum_{j \in N(i)} \kappa_{ij} \left( T_j^{(t)} - T_i^{(t)} \right)$
        \EndFor
        \State $\Delta T_{max} \leftarrow \max_{i} |T_i^{(t+1)} - T_i^{(t)}|$
        \State $t \leftarrow t + 1$
    \EndWhile
    
    \Statex \textit{\textcolor[RGB]{50,134,187}{// Step 3. Region Merging} }
    \State $\mathcal{R} \leftarrow$ Merge adjacent regions $(S_i, S_j)$ into a single semantic unit if $|T_i^{(\text{final})} - T_j^{(\text{final})}| < \tau_m$
    
    \State \textbf{Return} Final semantically complete region set $\mathcal{R} = \{R_1, \dots, R_M\}$
    \end{algorithmic}
    \end{algorithm}

\textit{\textbf{Final Iterative Update Rule.}}
Substituting Eq.~(\ref{eq:supp_temporal_discrete}) and Eq.~(\ref{eq:supp_spatial_discrete}) back into Eq.~(\ref{eq:supp_continuous_heat}), we obtain the heat conduction equation on the graph:
\begin{equation}
    \frac{T_i^{(t+1)} - T_i^{(t)}}{\Delta t} = \sum_{j \in N(i)} \kappa_{ij} \cdot \left(T_j^{(t)} - T_i^{(t)}\right)
\end{equation}
By rearranging the terms, we derive the final iterative update formula used in our Thermodiffusion-Guided Token Merging algorithm:
\begin{equation}
    T_i^{(t+1)} = T_i^{(t)} + \Delta t \sum_{j \in N(i)} \kappa_{ij} \left(T_j^{(t)} - T_i^{(t)}\right)
    \label{eq:supp_final_update}
\end{equation}
This equation mathematically demonstrates that in each step, the temperature (semantic feature) of a region is updated by absorbing information from its neighbors, weighted by their semantic affinity $\kappa_{ij}$, effectively simulating the physical process of thermal equilibrium to achieve boundary-preserving homogenization. 

To provide a clear understanding of our implementation details, we summarize the complete workflow of the Thermodiffusion-Guided Token Merging algorithm in Algorithm~\ref{alg:heat_token_merging}.

\begin{table*}[t]
    \centering
    \caption{List of hyperparameters for the Thermodiffusion-Guided Token Merging algorithm, including their symbols, optimal values, and physical interpretations.}
    \label{tab:hyperparameters}
    \renewcommand{\arraystretch}{1.3} 
    \small 
    \begin{tabularx}{\textwidth}{l|c|c|X}
    \toprule
    \textbf{Parameter} & \textbf{Symbol} & \textbf{Value} & \textbf{Physical Interpretation \& Impact} \\
    \midrule
    Base Diffusivity & $\kappa_0$ & 1.0 & Controls the global baseline speed of heat exchange between all adjacent regions. \\
    \midrule
    Color Sensitivity Scale & $\sigma_C$ & 5.0 & Controls sensitivity to color differences during diffusion. A larger value makes diffusion less sensitive to color variations, allowing heat to flow across minor spectral differences. \\
    \midrule
    Temperature Scale & $\sigma_T$ & 5.0 & Controls color sensitivity for initial temperature assignment ($T^{(0)}$). A smaller value requires high neighbor similarity to form a "Heat Source". \\
    \midrule
    Complexity Factor & $\alpha$ & 0.5 & Modulates the influence of texture complexity ($C_i$). A larger value accelerates heat exchange in high-frequency/complex regions. \\
    \midrule
    Merging Threshold & $\tau_m$ & 0.03 & The threshold for temperature difference. A larger value leads to more aggressive merging and coarser tokens. \\
    \bottomrule
    \end{tabularx}
    \end{table*}

    \begin{table*}[t]
    \centering
    \caption{Hyperparameter settings for different experimental ablation sets.}
    \label{tab:hyper_settings}

    \vspace{0.2cm}

    \setlength{\tabcolsep}{20pt}
    \renewcommand{\arraystretch}{1.15}
    \footnotesize
    \resizebox{\textwidth}{!}{%
    \begin{tabular}{l|ccccc}
    \toprule
    \textbf{Experiment Set} & \textbf{$\kappa_0$} & \textbf{$\sigma_C$} & \textbf{$\sigma_T$} & \textbf{$\alpha$} & \textbf{$\tau_m$} \\
    & (Diffusivity) & (Color Sens.) & (Temp. Scale) & (Complexity) & (Threshold) \\
    \midrule
    \textbf{Optimal (Ours)} & \textbf{1.0} & \textbf{5.0} & \textbf{5.0} & \textbf{0.5} & \textbf{0.03} \\
    \midrule
    Set A  & 2.0 & 10.0 & 10.0 & 0.5 & 0.03 \\
    Set B  & 1.0 & 5.0 & 5.0 & 0.0 & 0.03 \\
    Set C  & 4.0 & 5.0 & 5.0 & 0.5 & 0.03 \\
    \bottomrule
    \end{tabular}%
    }
    \end{table*}

\section{Hyperparameter Sensitivity Analysis}
\label{sec:hyper_analysis}

To validate the robustness of our proposed method and determine the optimal configuration for the thermodynamic parameters, we conducted a comprehensive hyperparameter sensitivity analysis. The Thermodiffusion-Guided Token Merging process is governed by five key parameters: base diffusivity ($\kappa_0$), color sensitivity ($\sigma_C$), temperature initialization scale ($\sigma_T$), complexity factor ($\alpha$), and merging threshold ($\tau_m$).

\subsection{Hyperparameter Definitions}

In this section, we provide a detailed analysis of the five hyperparameters involved in the Thermodiffusion-Guided Token Merging module. These parameters govern the initialization of the temperature field, the rate of semantic diffusion, and the final merging granularity. Table~\ref{tab:hyperparameters} summarizes the optimal values identified through empirical validation on the VRSBench and EarthVQA datasets.

\textit{\textbf{Initialization Parameter ($\sigma_T$).}}
The parameter $\sigma_T$ determines the "activation" of Heat Sources. 
\begin{itemize}
    \item \textbf{Impact:} A smaller $\sigma_T$ implies strict conditions for high initial temperatures; a region must be highly similar to its neighbors to be considered a core semantic object. Conversely, a larger $\sigma_T$ results in a broader distribution of high-temperature initial regions, leading to larger merged areas. We set $\sigma_T = 5.0$ to balance selectivity and coverage.
\end{itemize}

\textit{\textbf{Diffusion Dynamics Parameters ($\kappa_0, \sigma_C, \alpha$).}}
These parameters define the thermal diffusivity $\kappa_{ij}$, regulating how semantic information propagates.
\begin{itemize}
    \item \textbf{Base Diffusivity ($\kappa_0$):} Acts as a global scaling factor. A larger $\kappa_0$ accelerates the overall convergence speed but may lead to instability if the discrete time step $\Delta t$ is not adjusted accordingly.
    \item \textbf{Color Sensitivity ($\sigma_C$):} This parameter is critical for boundary preservation. A larger $\sigma_C$ (e.g., 5.0) indicates a higher tolerance (i.e., lower sensitivity) to color differences, allowing heat to flow rapidly even between regions with slight spectral discrepancies. This is essential for merging heterogeneous parts of a single object (e.g., a roof with shadows).
    \item \textbf{Complexity Factor ($\alpha$):} This parameter incorporates texture information. A higher $\alpha$ amplifies diffusivity in high-frequency regions (high gradient $C_i$), ensuring that complex textures are homogenized quickly, while smooth regions diffuse at a standard rate.
\end{itemize}

\textit{\textbf{Merging Parameter ($\tau_m$).}}
The final token generation is controlled by the merging threshold.
\begin{itemize}
\item \textbf{Merging Threshold ($\tau_m$):} This is the decision boundary for the final aggregation. A larger $\tau_m$ allows regions with larger temperature gradients to merge, resulting in fewer tokens. We set $\tau_m = 0.03$ to achieve object-level representation.
\end{itemize}

\subsection{Ablation Study on Thermodynamic Parameters}
\label{subsec:ablation_hyper_study}

To demonstrate the impact of each hyperparameter, we designed several comparative experimental sets by ablating the optimal parameters individually. The detailed settings are listed in Table~\ref{tab:hyper_settings}.

\begin{figure}[H]
    \centering
    \includegraphics[width=\columnwidth]{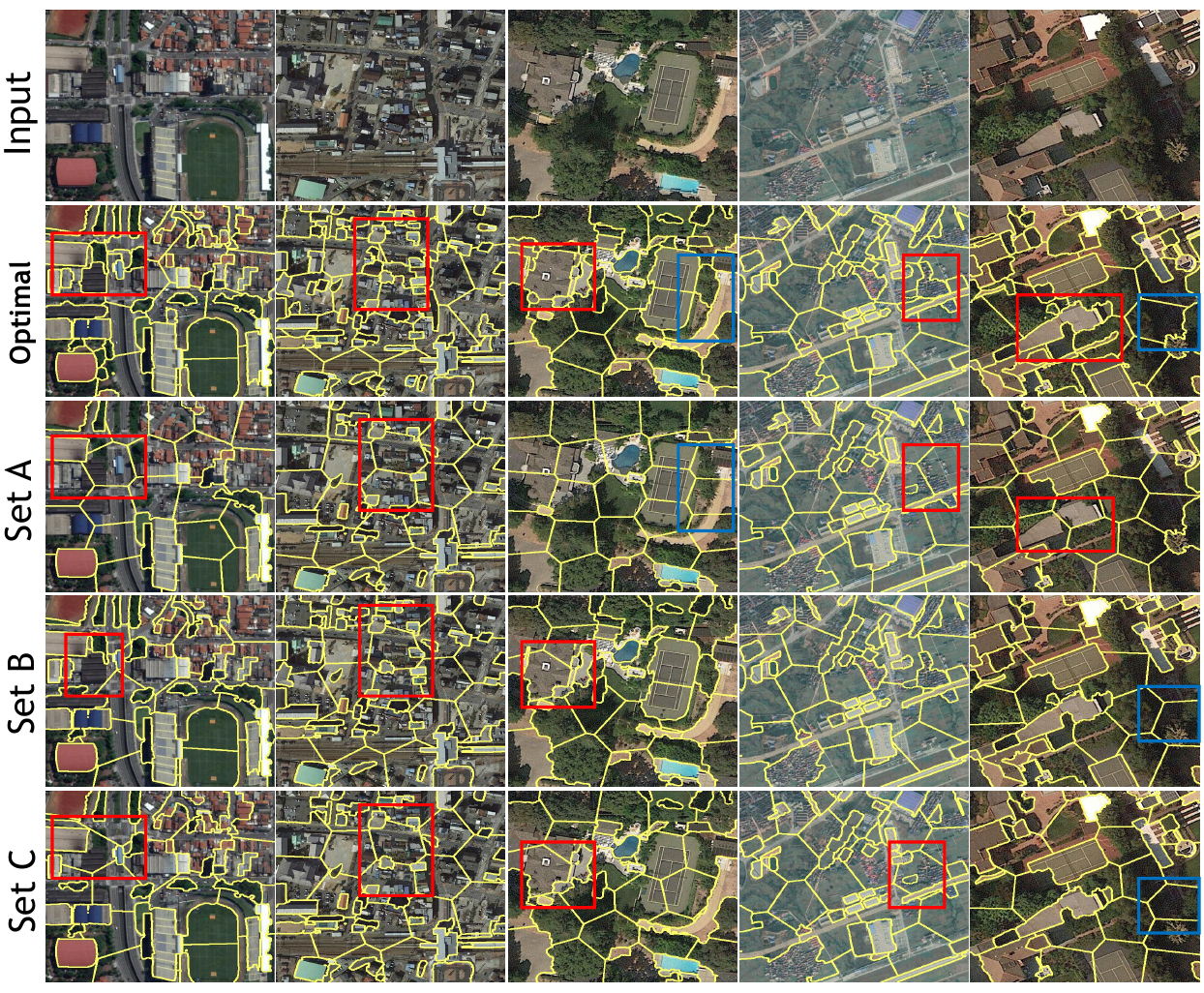}
    \caption{Visual comparison of tokenization results under different hyperparameter configurations.}
    \label{fig:hyper_vis}
    \vspace{-0.2cm}
\end{figure}

\textit{\textbf{Experimental Setup.}} 
We established three comparative groups (Set A, B, C) alongside our final Optimal configuration.
\begin{itemize}
    \item \textbf{Optimal (Ours):} The balanced configuration ($\kappa_0=1.0, \sigma_C=5.0, \sigma_T=5.0, \alpha=0.5, \tau_m=0.03$) used in our main experiments.
    \item \textbf{Set A:} We simultaneously increased the base diffusivity ($\kappa_0=2.0$), color sensitivity ($\sigma_C=10.0$), and temperature scale ($\sigma_T=10.0$). This tests the model's behavior under conditions that favor broader initial heat activation and more aggressive region aggregation.
    \item \textbf{Set B:} We set the complexity factor to $\alpha=0.0$, effectively disabling the texture-aware diffusivity modulation.
    \item \textbf{Set C:} We significantly increased the base diffusivity to $\kappa_0=4.0$ to evaluate the stability of the heat equation solver under extreme high-speed diffusion.
\end{itemize}

\textit{\textbf{Qualitative Analysis.}}
The visual comparison of tokenization results under these settings is presented in Figure~\ref{fig:hyper_vis}.
\begin{itemize}
    \item \textbf{Under-segmentation in Set A:} When the parameters for diffusion, color tolerance, and temperature activation are excessively high, as in Set A, the algorithm leads to severe under-segmentation. Small but semantically distinct objects (e.g., small buildings surrounding the stadium) are incorrectly merged into the background, causing the MLLM to lose fine-grained details.

    \item \textbf{Boundary Noise in Set B:} By removing the texture complexity constraint ($\alpha=0.0$), Set B fails to accurately delineate boundaries in texture-rich areas. The generated tokens become irregular and fail to align with inherent high-frequency texture transitions, confirming that texture-guided diffusion is crucial for handling the complex heterogeneity of remote sensing imagery.

    \item \textbf{Over-smoothing in Set C:} The extremely high base diffusivity ($\kappa_0=4.0$) in Set C causes semantic heat to propagate too rapidly, eroding fine-grained details before stable equilibrium boundaries can be established. Consequently, this leads to severe boundary degradation and a significant loss of morphological fidelity in the generated tokens.
    
    \item \textbf{Superiority of Optimal Settings:} In contrast, our Optimal configuration achieves the best balance. It successfully preserves the semantic integrity of large objects while accurately separating adjacent distinct regions. The generated tokens exhibit high boundary adherence and appropriate granularity, providing a solid foundation for the subsequent MLLM visual understanding.
\end{itemize}

\begin{table*}[t]
    \centering
    \caption{Ablation on region proposal dependency on \textbf{VRSBench}~\cite{li2024vrsbench}. \textit{w/o Thermo} indicates using raw proposals without our aggregation module. \textit{Frontend Params} strictly includes the 24.0M parameters of the thermodiffusion module where applicable. \textit{All} represents the sample-weighted average accuracy. \textcolor{red}{Red} indicates the best performance, while \textcolor{blue}{blue} denotes the second best.}
    \label{tab:ablation_vrsbench}
    
    \setlength{\tabcolsep}{3.5pt}
    \renewcommand{\arraystretch}{1.2}
    \small
    
    \resizebox{\textwidth}{!}{%
    \begin{tabular}{l | c c | cccccccccc|c}
    \toprule
    \textbf{Proposal Generator} & \textbf{\makecell{Frontend \\ Latency (ms)}} & \textbf{\makecell{Frontend \\ Params (M)}} & \textbf{Category} & \textbf{Presence} & \textbf{Quantity} & \textbf{Color} & \textbf{Shape} & \textbf{Size} & \textbf{Position} & \textbf{Direction} & \textbf{Scene} & \textbf{Reasoning} & \textbf{All} \\
    \midrule
    
    FastSAM (w/o Thermo) & \textcolor{red}{12.1} & \textcolor{blue}{68.0} & 76.53 & 85.27 & 51.56 & 68.78 & 71.82 & 58.55 & 59.88 & 50.52 & 78.59 & 68.74 & 69.84 \\
    
    SLIC + Thermo & 45.2 & \textcolor{red}{24.0} & 82.52 & 90.13 & 55.21 & 75.84 & 76.56 & 62.48 & 65.47 & 53.44 & 82.37 & 72.84 &  74.77\\
    
    SAM-H + Thermo & 1468.4 & 624.0 & \textcolor{blue}{84.23} & \textcolor{red}{91.61} & \textcolor{blue}{56.12} & \textcolor{red}{77.49} & \textcolor{red}{78.15} & \textcolor{red}{64.53} & \textcolor{blue}{67.52} & \textcolor{blue}{55.03} & \textcolor{blue}{83.54} & \textcolor{red}{74.57} & \textcolor{blue}{76.28} \\
    
    \rowcolor[RGB]{224,255,255} FastSAM + Thermo (Ours) & \textcolor{blue}{15.3} & 92.0 & \textcolor{red}{84.45} & \textcolor{blue}{91.41} & \textcolor{red}{56.36} & \textcolor{blue}{77.32} & \textcolor{blue}{78.07} & \textcolor{blue}{64.20} & \textcolor{red}{67.96} & \textcolor{red}{55.35} & \textcolor{red}{83.75} & \textcolor{blue}{74.28} & \textcolor{red}{76.37} \\
    
    \bottomrule
    \end{tabular}%
    }
\end{table*}

\begin{table*}[t]
    \centering
    \caption{Ablation on region proposal dependency on \textbf{EarthVQA}~\cite{wang2024earthvqa}, showing that gains stem from thermodiffusion rather than a specific frontend.}
    \label{tab:ablation_earthvqa}
    
    \setlength{\tabcolsep}{10pt}
    \renewcommand{\arraystretch}{1.2}
    \small
    
    \resizebox{\textwidth}{!}{%
    \begin{tabular}{l | c c | cccccc|c}
    \toprule
    \textbf{Proposal Generator} & \textbf{\makecell{Frontend \\ Latency (ms)}} & \textbf{\makecell{Frontend \\ Params (M)}} & \textbf{BasJu} & \textbf{RelJu} & \textbf{BasCo} & \textbf{RelCo} & \textbf{ObjAn} & \textbf{ComAn} & \textbf{All} \\
    \midrule
    
    FastSAM (w/o Thermo) & \textcolor{red}{12.1} & \textcolor{blue}{68.0} & 75.13 & 77.28 & 68.32 & 74.56 & 55.09 & 50.43 & 70.60 \\
    
    SLIC + Thermo & 45.2 & \textcolor{red}{24.0} & 81.12 & 83.96 & 73.25 & 79.83 & 65.17 & 56.48 & 76.84 \\
    
    SAM-H + Thermo & 1468.4 & 624.0 & \textcolor{red}{82.83} & \textcolor{blue}{85.19} & \textcolor{blue}{74.30} & \textcolor{red}{82.04} & \textcolor{blue}{67.81} & \textcolor{blue}{59.06} & \textcolor{blue}{78.39} \\
    
    \rowcolor[RGB]{224,255,255} FastSAM + Thermo (Ours) & \textcolor{blue}{15.3} & 92.0 & \textcolor{blue}{82.60} & \textcolor{red}{85.94} & \textcolor{red}{75.06} & \textcolor{blue}{81.67} & \textcolor{red}{68.87} & \textcolor{red}{59.49} & \textcolor{red}{78.88} \\
    
    \bottomrule
    \end{tabular}%
    }
\end{table*}

\section{Ablation on Region Proposal Dependency}

To demonstrate that the performance gains of \name{} stem primarily from our proposed thermodiffusion aggregation rather than a heavy reliance on a specific region proposal model, we substitute FastSAM with alternative proposal generators: SLIC~\cite{achanta2012slic} and SAM-H~\cite{kirillov2023segment}. For a strictly fair comparison, the 24.0M parameter overhead of the thermodiffusion module is explicitly added to the frontend parameters for all relevant configurations. The ablation results across the sub-categories of VRSBench and EarthVQA are detailed in Table~\ref{tab:ablation_vrsbench} and Table~\ref{tab:ablation_earthvqa}, respectively.

When using raw proposals generated by FastSAM without the thermodiffusion module (w/o Thermo.), the overall performance drops drastically (e.g., to 69.84{\%} on VRSBench and 70.60{\%} on EarthVQA). This confirms that relying solely on fine-grained segmentation outputs leads to semantic fragmentation and fails to provide coherent object-level tokens for the MLLM. Applying thermodiffusion aggregation yields significant accuracy improvements across all proposal generators, demonstrating that the primary performance gain is driven by the thermodiffusion aggregation module.

Notably, substituting FastSAM with the heavy-weight SAM-H yields highly competitive performance, slightly outperforming our method on specific fine-grained subsets due to its massive zero-shot priors. However, \name{} reclaims the lead in the overall sample-weighted accuracy on both datasets (76.37{\%} vs. 76.28{\%} on VRSBench; 78.88{\%} vs. 78.39{\%} on EarthVQA). More importantly, the grid-prompting mechanism of SAM-H in "Everything mode" incurs a catastrophic frontend latency (1468.4 ms) and a bloated parameter overhead (624.0M), completely bottlenecking the pipeline's efficiency. Conversely, while SLIC offers a lightweight alternative, it suffers in boundary precision and CPU-bound latency (45.2 ms), leading to a performance drop. These results comprehensively prove that \name{} is fundamentally agnostic to the initial proposal generator, and FastSAM combined with thermodiffusion offers the absolute optimal trade-off among frontend latency (15.3 ms), parameter efficiency (92.0M), and downstream reasoning accuracy.

\section{Generalization to Fundamental Vision Tasks}
\label{sec:appendix-vision-tasks}

To further validate the generalization capability of \name{} beyond VQA tasks, we extend our evaluation to fundamental vision tasks including visual grounding and semantic segmentation.

\subsection{Visual Grounding}

\begin{figure*}[t]
\centering
\includegraphics[width=1\textwidth]{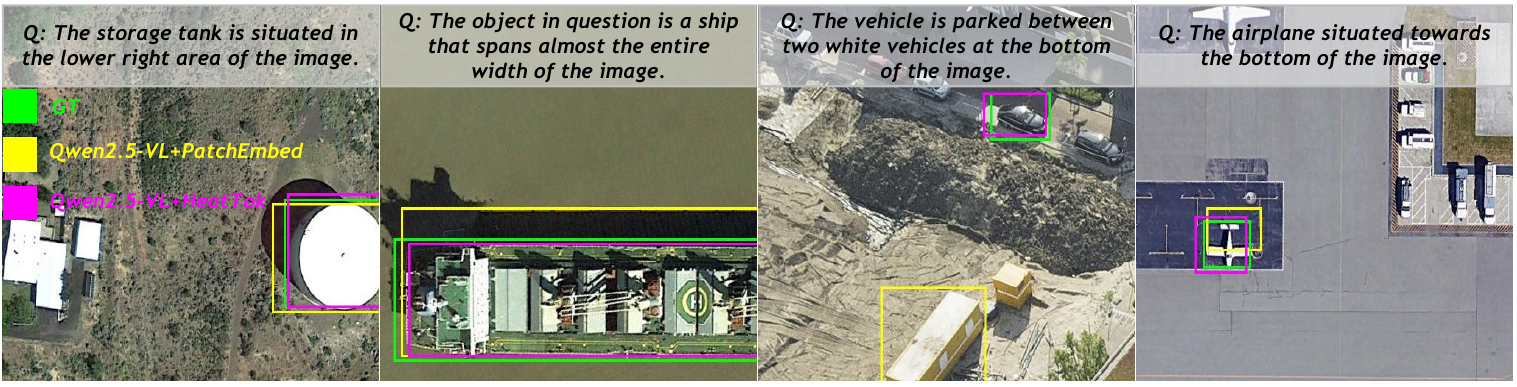}
\vspace{-0.3cm}
\caption{Visual grounding comparison between Patch Embed and \name{} on Qwen2.5-VL, highlighting improved object-boundary alignment and more precise spatial localization.}
\label{fig:visual-grounding}
\vspace{-0.2cm}
\end{figure*}

Figure~\ref{fig:visual-grounding} presents qualitative visual grounding results on VRSBench. Compared to the standard Patch Embed tokenization, \name{} produces more accurate bounding box predictions that are better aligned with object boundaries. This demonstrates that our thermodiffusion-based tokenization effectively preserves the geometric integrity of objects, enabling more precise spatial localization.

\subsection{Semantic Segmentation}

\begin{figure*}[t]
\centering
\includegraphics[width=1\textwidth,height=4cm,keepaspectratio=false]{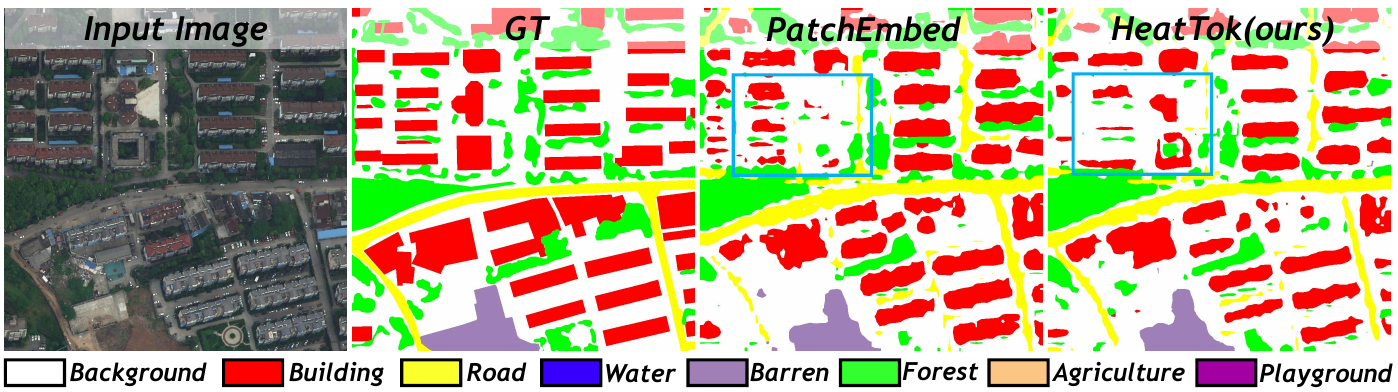}
\vspace{-0.3cm}
\caption{Qualitative segmentation results on EarthVQA, showing more accurate boundaries and better preserved land-cover integrity.}
\label{fig:segmentation}
\vspace{-0.2cm}
\end{figure*}

Figure~\ref{fig:segmentation} visualizes the segmentation results on EarthVQA. \name{} produces more accurate boundaries and better preserves the integrity of land-cover regions, demonstrating that our thermodiffusion-based tokenization extends effectively to dense prediction tasks.

\begin{figure*}[t]
    \centering
    \includegraphics[width=1\textwidth]{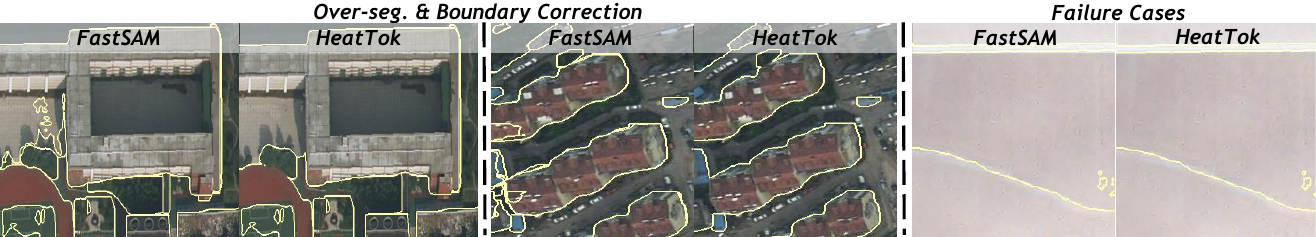}
    \vspace{-0.3cm}
    \caption{Robustness analysis. \name{} corrects over-segmentation by merging fragmented proposals into coherent regions (left), but performance degrades under severe haze (right).}
    \label{fig:robustness}
    \vspace{-0.2cm}
\end{figure*}

\section{Additional Qualitative Results}

\subsection{Tokenization Visualization}

\begin{figure*}[t]
\centering
\includegraphics[width=1\textwidth]{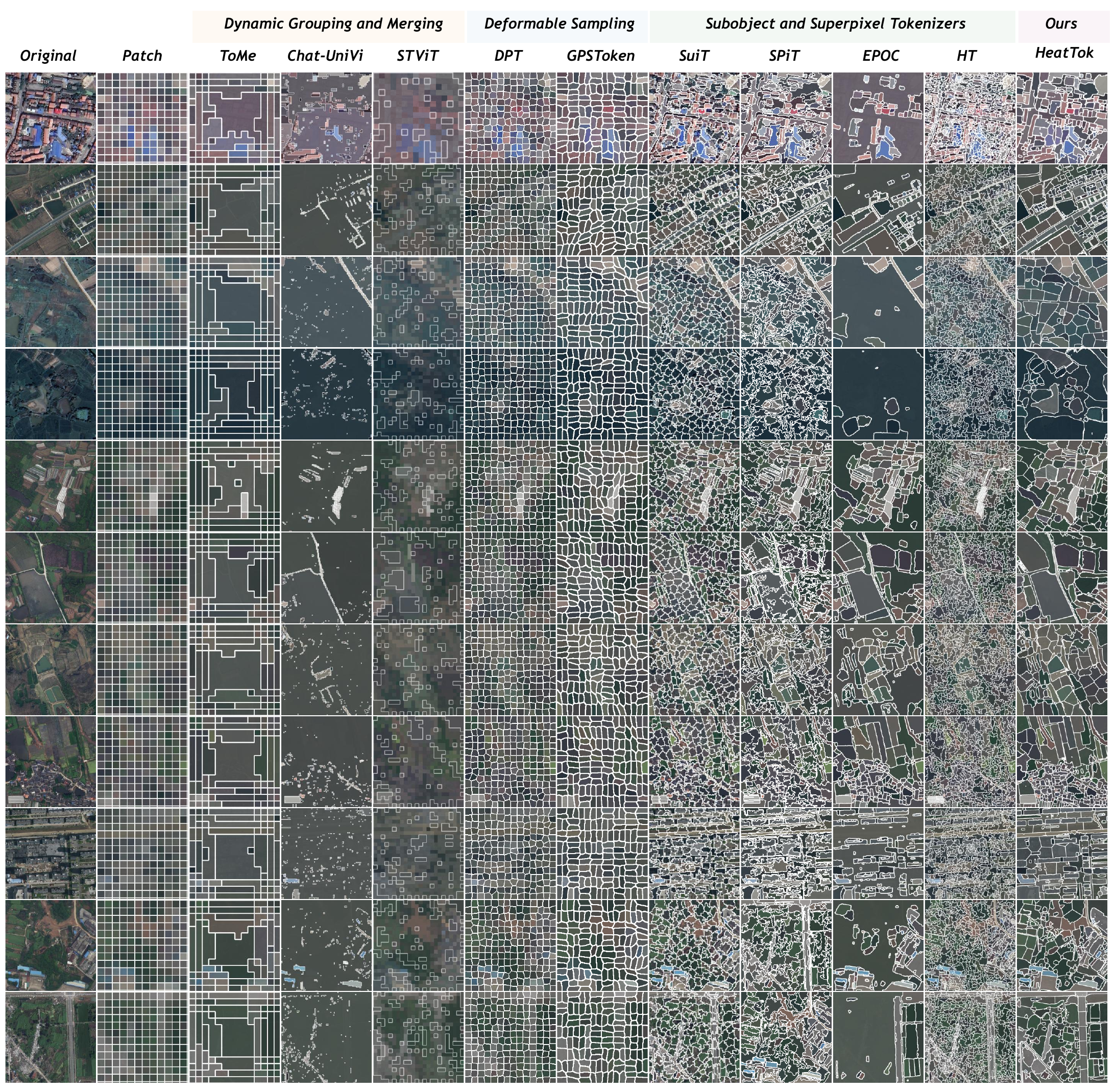}
\vspace{-0.6cm}
\caption{Visual comparisons of tokenization results across different paradigms. Compared to baseline methods, \name{} successfully preserves crisp object boundaries, maintains geometric continuity, and effectively avoids over-segmentation.}
\label{fig:vis_comparison2}
\end{figure*}

To further validate the robustness and generalization ability of our proposed method, we provide additional visualization comparisons on randomly selected samples from the evaluation set. As shown in Fig.~\ref{fig:vis_comparison2}, we comprehensively compare \name{} against other representative tokenization paradigms. These visualizations intuitively demonstrate that our thermodiffusion-based approach can adaptively group semantically consistent regions, preserving the structural integrity of diverse geographical objects while avoiding the severe over-segmentation and rigid geometric cuts seen in previous methods.

\subsection{Case Study}

In this section, we present additional remote sensing visual question answering (VQA) case studies to qualitatively evaluate the direct impact of different tokenization strategies on the reasoning capabilities of the Multimodal Large Language Model (MLLM). 

Table~\ref{case2} and Table~\ref{case3} detail the visual question answering results alongside the corresponding tokenization visualizations for two distinct geographical scenarios (i.e., curved park layouts and meandering rivers). These cases further corroborate that the object-level semantic integrity preserved by \name{} is crucial for accurate multi-modal grounding and robust reasoning.

\begin{table*}[ht]
\centering
\vspace{-0.2cm}
\caption{Case study on remote sensing visual question answering. We show the visual question answering results of a specific scenario along with the tokenization visualizations for different methods. \textcolor{red}{Red} indicates hallucinations or incorrect perceptions; \textcolor{ForestGreen}{Green} indicates accurate alignment with the ground truth.}
\label{case2}
\small
\begin{tabularx}{\linewidth}{X}
\toprule
\textbf{Remote Sensing Image \& Tokenization Visualization:} \\[2pt]
\begin{center}
\includegraphics[width=1\linewidth]{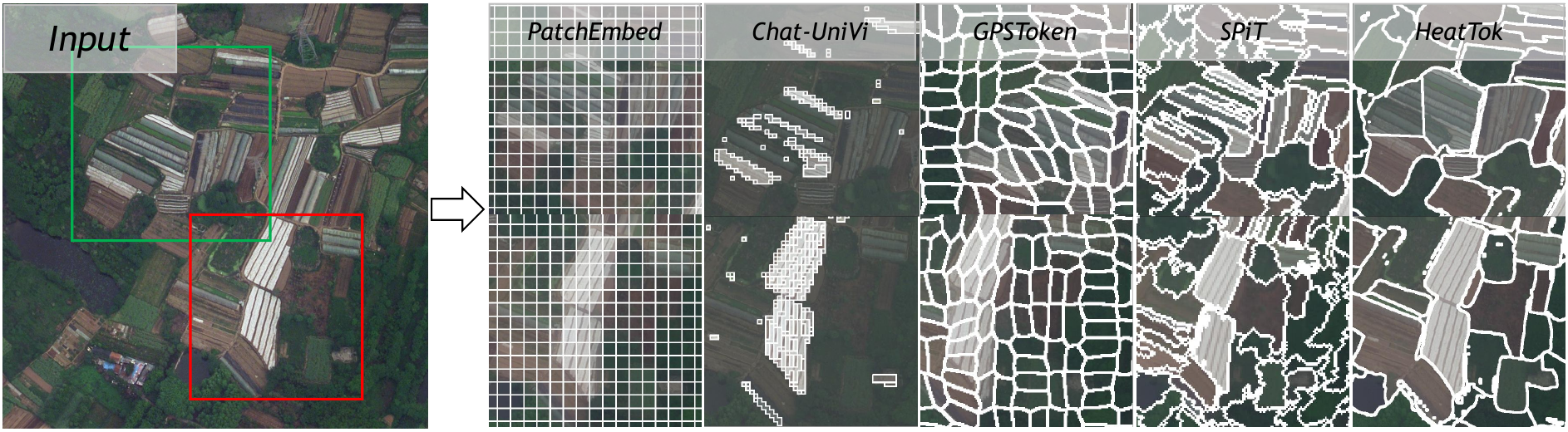}
\end{center} \\

\midrule
\textbf{Prompt:} What are the land use types in this scene? What is the situation of barren land? \\

\midrule
\textbf{Ground Truth:} There are \textcolor{ForestGreen}{park, and agricultural areas}. \textcolor{ForestGreen}{There is no barren land}.\\

\midrule
\textit{\textbf{Patch Embed:}} I can observe that there are \textcolor{ForestGreen}{park, and agricultural areas} within the scene. However, it seems the \textcolor{red}{barren land is scattered} across multiple small patches. \vspace{2pt}\newline
\textit{\textbf{Analysis:}} The uniform grid \textcolor{blue}{rigidly cuts across the curved boundaries} of the park and fields. This \textcolor{blue}{destruction of geometric continuity} causes the model to misinterpret local texture variations (like winding paths or dry grass) as scattered barren land. \\

\midrule
\multicolumn{1}{c}{\textit{\textbf{Dynamic Grouping and Merging}}} \\
\textit{\textbf{ToMe:}} I can identify \textcolor{red}{industrial} zones alongside \textcolor{ForestGreen}{agricultural areas}. The \textcolor{red}{barren land appears to be distributed} unevenly. \\[2pt]
\textit{\textbf{Chat-UniVi:}} Based on the image, there are \textcolor{ForestGreen}{park}, \textcolor{red}{residential}, and \textcolor{ForestGreen}{agricultural areas}. Fortunately, \textcolor{ForestGreen}{there is no barren land} visible in this specific environment. \\[2pt]
\textit{\textbf{Quadtree:}} The land use types predominantly include \textcolor{red}{residential} and \textcolor{ForestGreen}{agricultural areas}. Additionally, the \textcolor{red}{barren land is concentrated} in the lower section of the region. \\[2pt]
\textit{\textbf{HOOK:}} The scene is mostly \textcolor{red}{residential} and \textcolor{red}{commercial}. A large patch of \textcolor{red}{barren land dominates} the upper left corner. \\[2pt]
\textit{\textbf{STViT:}} There are \textcolor{ForestGreen}{park} areas mixed with \textcolor{red}{dense residential buildings}. Some \textcolor{red}{scattered barren land} is present near the edges. \\[2pt]
\textit{\textbf{Analysis:}} Although these methods attempt to group adjacent tokens, the Chat-UniVi visualization reveals that they remain \textcolor{blue}{inherently constrained by rigid rectangular primitives}. This inflexible spatial quantization is inadequate for delineating the continuous curved contours of the park, ultimately misleading the MLLM to \textcolor{blue}{hallucinate artificial residential structures} and non-existent barren regions. \\

\midrule
\multicolumn{1}{c}{\textit{\textbf{Deformable Sampling}}} \\
\textit{\textbf{DPT:}} The area is entirely a \textcolor{red}{dense urban residential} zone. A vast stretch of \textcolor{red}{barren land covers the majority} of the visible space. \vspace{2pt}\newline
\textit{\textbf{GPSToken:}} The land use types are primarily \textcolor{red}{commercial} and \textcolor{ForestGreen}{agricultural}. Furthermore, the \textcolor{red}{barren land is clearly located in the center} of the complex. \vspace{2pt}\newline
\textit{\textbf{Analysis:}} The GPSToken visualization demonstrates that deformable points focus heavily on local high-frequency details but \textcolor{blue}{severely distort the macro-level layout}. Because the MLLM \textcolor{blue}{loses the global topological shape} of the fields, it confidently hallucinates "commercial" areas and a centralized barren patch. \\

\midrule
\multicolumn{1}{c}{\textit{\textbf{Subobject and Superpixel Tokenizers}}}\\
\textit{\textbf{SuiT:}} There are \textcolor{ForestGreen}{parks} and \textcolor{red}{industrial facilities}. The \textcolor{red}{barren land is adjacent} to the main roads. \vspace{2pt}\newline
\textit{\textbf{SPiT:}} The land use types in this image include \textcolor{ForestGreen}{park}, \textcolor{red}{residential}, and \textcolor{ForestGreen}{agricultural areas}. Moreover, the \textcolor{red}{barren land is scattered} randomly throughout the site. \vspace{2pt}\newline
\textit{\textbf{EPOC:}} I observe \textcolor{ForestGreen}{agricultural fields} and \textcolor{red}{commercial buildings}. Small pockets of \textcolor{red}{barren land are scattered} widely. \vspace{2pt}\newline
\textit{\textbf{$\partial$HT:}} There are \textcolor{ForestGreen}{park} areas and \textcolor{red}{residential blocks}. \textcolor{ForestGreen}{There is no barren land} present here. \vspace{2pt}\newline
\textit{\textbf{Analysis:}} Superpixel algorithms like SPiT \textcolor{blue}{excessively fragment} the continuous grassy regions into erratic, jagged pieces. This \textcolor{blue}{over-segmentation noise} breaks the visual consistency of the park's landscaping, tricking the model into identifying these irregular fragments as scattered barren land and residential zones. \\

\midrule
\textit{\textbf{HeatTok (Ours):}} The dominant land use types are clearly well-maintained \textcolor{ForestGreen}{park, and agricultural areas}. \textcolor{ForestGreen}{There is no barren land} anywhere in the image. \vspace{2pt}\newline
\textit{\textbf{Analysis:}} By utilizing thermodiffusion aggregation, \name{} \textcolor{blue}{elegantly preserves the continuous, sweeping curves} of the park layout and the intact agricultural fields. Maintaining this \textcolor{blue}{precise geometric and semantic integrity} allows the MLLM to easily recognize the correct land use categories and firmly conclude the absence of barren land \textcolor{blue}{without any hallucinations}. \\

\bottomrule
\end{tabularx}
\end{table*}

\begin{table*}[ht]
\centering
\vspace{-0.2cm}
\caption{Case study on remote sensing visual question answering. We show the visual question answering results of a specific scenario along with the tokenization visualizations for different methods. \textcolor{red}{Red} indicates hallucinations or incorrect perceptions; \textcolor{ForestGreen}{Green} indicates accurate alignment with the ground truth.}
\label{case3}
\small
\begin{tabularx}{\linewidth}{X}
\toprule
\textbf{Remote Sensing Image \& Tokenization Visualization:} \\[2pt]
\begin{center}
\includegraphics[width=1\linewidth]{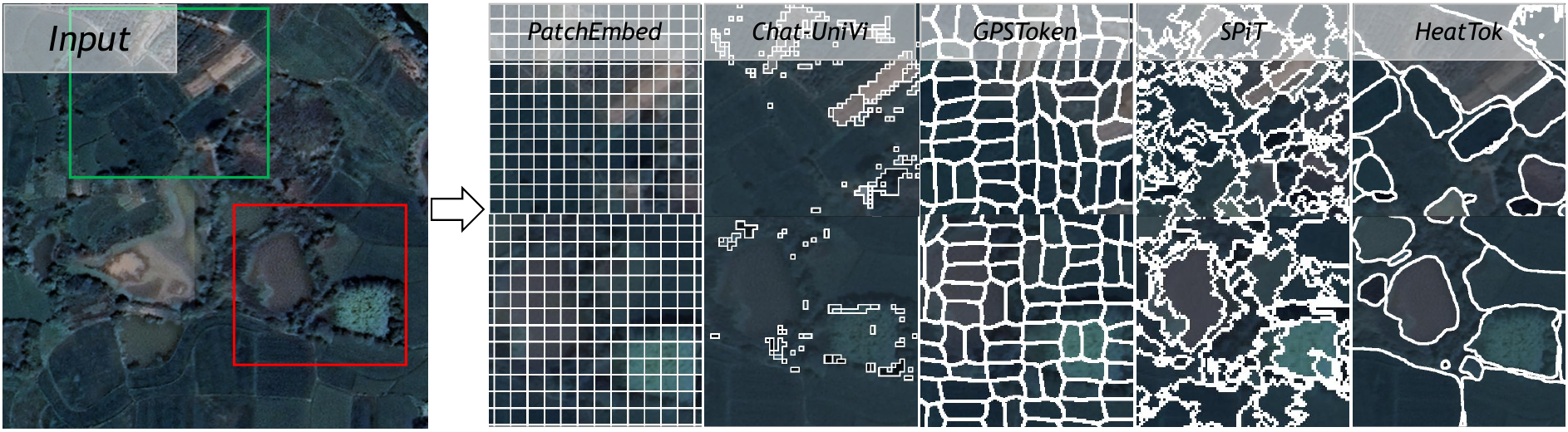}
\end{center} \\

\midrule
\textbf{Prompt:} What are the water situations around the agricultural land? What are the needs for the renovation of villages? \\

\midrule
\textbf{Ground Truth:} There is \textcolor{ForestGreen}{clean water} near the agriculture land. The \textcolor{ForestGreen}{roads} need to be \textcolor{ForestGreen}{improved}. \\

\midrule
\textit{\textbf{Patch Embed:}} Based on the scene, I can see \textcolor{red}{dried-up water channels} near the agriculture land. The \textcolor{ForestGreen}{roads} need to be \textcolor{ForestGreen}{improved}, and \textcolor{red}{new residential buildings should be constructed}. \vspace{2pt}\newline
\textit{\textbf{Analysis:}} The uniform grid partitioning \textcolor{blue}{rigidly bisects the meandering river} and adjacent fields. This \textcolor{blue}{spatial fragmentation} disrupts the clear water textures, causing the model to misinterpret the disjointed river segments as "dried-up channels" and hallucinate unrelated construction needs. \\

\midrule
\multicolumn{1}{c}{\textit{\textbf{Dynamic Grouping and Merging}}} \\
\textit{\textbf{ToMe:}} There are \textcolor{red}{polluted streams} near the agriculture land. The \textcolor{ForestGreen}{roads} need to be \textcolor{ForestGreen}{improved} and \textcolor{red}{residential areas expanded}. \\[2pt]
\textit{\textbf{Chat-UniVi:}} There appear to be \textcolor{red}{scattered muddy pools} near the fields. The narrow dirt \textcolor{ForestGreen}{roads} need to be \textcolor{ForestGreen}{improved}. \\[2pt]
\textit{\textbf{Quadtree:}} The area seems to \textcolor{red}{lack prominent water features} near the fields. \textcolor{red}{Infrastructure} improvements and \textcolor{red}{housing} upgrades are critically needed. \\[2pt]
\textit{\textbf{HOOK:}} The farmland is completely \textcolor{red}{devoid of nearby water resources}. There are \textcolor{red}{no specific needs} for village renovation visible in this particular area. \\[2pt]
\textit{\textbf{STViT:}} Some \textcolor{red}{stagnant water} is visible. The \textcolor{ForestGreen}{roadways} need to be \textcolor{ForestGreen}{improved}. \\[2pt]
\textit{\textbf{Analysis:}} Chat-UniVi's visualization shows these methods remain \textcolor{blue}{constrained by rigid rectangular primitives}. This \textcolor{blue}{inflexible block-wise quantization} fails to delineate the river's continuous meandering trajectory. Consequently, the fragmented semantics mislead the MLLM into misclassifying the water as "muddy pools," triggering cascaded hallucinations of unwarranted housing upgrades. \\

\midrule
\multicolumn{1}{c}{\textit{\textbf{Deformable Sampling}}} \\
\textit{\textbf{DPT:}} The area is \textcolor{red}{completely dry} without any rivers. The village needs a \textcolor{red}{completely new water supply system} due to the arid environment. \vspace{2pt}\newline
\textit{\textbf{GPSToken:}} I \textcolor{red}{cannot detect any water} area around the agricultural land. The agricultural land needs to be \textcolor{red}{restructured} and the \textcolor{red}{buildings upgraded}. \vspace{2pt}\newline
\textit{\textbf{Analysis:}} The GPSToken visualization illustrates that deformable sampling produces \textcolor{blue}{dense but spatially fragmented focal points} that latch onto arbitrary local textures. By \textcolor{blue}{discarding the macro-level topological continuity} of the river, the model completely fails to perceive the water body, hallucinating a dry environment and inventing the need for water supply systems. \\

\midrule
\multicolumn{1}{c}{\textit{\textbf{Subobject and Superpixel Tokenizers}}}\\
\textit{\textbf{SuiT:}} There are \textcolor{red}{highly polluted} water bodies adjacent to the farm plots. There is a need to \textcolor{ForestGreen}{improve} the dirt \textcolor{ForestGreen}{roads} and \textcolor{red}{build bridges}. \vspace{2pt}\newline
\textit{\textbf{SPiT:}} I observe \textcolor{red}{fragmented muddy water} around the agricultural land. The \textcolor{ForestGreen}{roads} need \textcolor{ForestGreen}{improvement}. \vspace{2pt}\newline
\textit{\textbf{EPOC:}} I see only \textcolor{red}{dry riverbeds} in the area. The \textcolor{ForestGreen}{roads} need to be \textcolor{ForestGreen}{improved} for better rural access. \vspace{2pt}\newline
\textit{\textbf{$\partial$HT:}} Yes, \textcolor{red}{contaminated ponds} are located nearby. The \textcolor{ForestGreen}{roads} need to be \textcolor{ForestGreen}{improved}. \vspace{2pt}\newline
\textit{\textbf{Analysis:}} Superpixel algorithms like SPiT rely heavily on low-level pixel variations, leading to \textcolor{blue}{excessive over-segmentation} of the river surface. Breaking the homogeneous water area into \textcolor{blue}{fractured, jagged fragments} introduces visual noise, tricking the model into classifying it as "fragmented muddy water" and fabricating extra renovation tasks. \\

\midrule
\textit{\textbf{HeatTok (Ours):}} I can clearly observe that \textcolor{ForestGreen}{clean water} are flowing near the agriculture land. For the renovation of the village, the dirt \textcolor{ForestGreen}{roads} need to be \textcolor{ForestGreen}{improved} to facilitate better transportation. \vspace{2pt}\newline
\textit{\textbf{Analysis:}} By leveraging thermodiffusion aggregation, \name{} \textcolor{blue}{precisely delineates the continuous, winding geometry} of the river and the road networks. Maintaining this \textcolor{blue}{object-level semantic integrity} enables the MLLM to correctly assess the "clean" state of the water and accurately pinpoint the road improvement needs \textcolor{blue}{without any hallucinations}. \\

\bottomrule
\end{tabularx}
\end{table*}

\section{Limitations and Future Work}

While \name{} achieves consistent improvements across diverse RS benchmarks, we acknowledge two main limitations regarding its broader applicability.

\subsection{Generalization to Natural Images}
\label{sec:applicability}

Our framework is fundamentally motivated by the unique properties of RS imagery, which features extensive homogeneous regions and multi-scale, irregular geometric structures. The thermodiffusion mechanism explicitly addresses these characteristics by aggregating irregular geometries into coherent, well-bounded region-level units. While the formulation is inherently task- and domain-agnostic, standard natural images exhibit distinct object-centric characteristics. We do not presume uniform performance gains across general natural image benchmarks; extending this framework to high-resolution natural imagery remains a direction for future research.

\subsection{Robustness and Failure Cases}
\label{sec:robustness}

Our ablation studies confirm that performance gains stem from the thermodiffusion-based aggregation mechanism rather than any specific region proposal method. As shown in Figure~\ref{fig:robustness}, thermodiffusion corrects over-segmentation and erroneous boundaries by merging fragmented proposals into coherent regions, demonstrating robustness to initial segmentation errors in complex RS scenes. However, under extremely degraded imaging conditions (e.g., severe haze or heavy fog), performance is constrained by frontend visual priors—compromised region proposals may limit overall system performance. Future work will explore domain-adaptive perception mechanisms tailored for such extreme scenarios.

\clearpage
\putbib
\end{bibunit}

\end{document}